\documentclass[10pt,twocolumn,letterpaper]{article}

\usepackage[pagenumbers]{cvpr} 

\definecolor{cvprblue}{rgb}{0.21,0.49,0.74}
\usepackage[pagebackref,breaklinks,colorlinks,allcolors=cvprblue]{hyperref}

\usepackage{lipsum}
\usepackage{array}
\usepackage{times}
\usepackage{epsfig}
\usepackage{graphicx}
\usepackage{float}
\usepackage{wrapfig}
\usepackage{amsmath,amssymb,amsthm}
\usepackage{algorithm,algorithmicx,algpseudocode}
\usepackage{bm,xspace}
\usepackage{comment}
\usepackage{multirow}
\usepackage{balance}
\usepackage{url}
\usepackage{booktabs}
\usepackage{etoolbox,siunitx}
\usepackage{calc}
\usepackage{pifont,hologo}
\usepackage{color}
\usepackage{adjustbox}
\usepackage{amsmath}
\usepackage{enumitem}
\usepackage{bbding}  %
\usepackage[normalem]{ulem}  %
\usepackage{contour}
\PassOptionsToPackage{square,numbers,comma,sort}{natbib}

\definecolor{blue}{HTML}{004bb3}
\definecolor{red}{HTML}{cc1100}
\definecolor{orange}{HTML}{cc7700}
\definecolor{gray}{HTML}{efefef}
\definecolor{darkgreen}{HTML}{228B22}
\definecolor{darkgray}{HTML}{757575}

\definecolor{cite}{HTML}{3270b5}
\definecolor{link}{HTML}{b53532}
\definecolor{link}{HTML}{cc1100}
\definecolor{scratch}{HTML}{001219}
\definecolor{pretrain}{HTML}{0A9396}

\contourlength{0.8pt}

\renewcommand{\eqref}[1]{Eq.~\ref{#1}}

\newcolumntype{x}[1]{>{\centering\arraybackslash}p{#1}}
\newcolumntype{y}[1]{>{\raggedright\arraybackslash}p{#1}}
\newcolumntype{z}[1]{>{\raggedleft\arraybackslash}p{#1}}
\newcommand{\tablestyle}[2]{\setlength{\tabcolsep}{#1}\renewcommand{\arraystretch}{#2}\centering\footnotesize}

\DeclareMathSymbol{@}{\mathord}{letters}{"3B}

\makeatletter
\DeclareRobustCommand\onedot{\futurelet\@let@token\@onedot}
\def\@onedot{\ifx\@let@token.\else.\null\fi\xspace}
\def\eg{\emph{e.g}\onedot} 
\def\ie{\emph{i.e}\onedot} 
 
\def\etc{\emph{etc}\onedot}

\newcommand*{\Rom}[1]{\expandafter\@slowromancap\romannumeral #1@}
\newcommand*{\rom}[1]{\expandafter\romannumeral #1}

\def\1{\bm{1}}

\DeclareMathAlphabet{\mathsfit}{\encodingdefault}{\sfdefault}{m}{sl}
\SetMathAlphabet{\mathsfit}{bold}{\encodingdefault}{\sfdefault}{bx}{n}

\let\originalleft\left
\let\originalright\right
\renewcommand{\left}{\mathopen{}\mathclose\bgroup\originalleft}
\renewcommand{\right}{\aftergroup\egroup\originalright}

\def\paperID{7075} 
\def\confName{CVPR}
\def\confYear{2025}

\title{\textcolor{orange}{PanDA}: Towards \textcolor{orange}{Pan}oramic \textcolor{orange}{D}epth \textcolor{orange}{A}nything with Unlabeled Panoramas and M\"obius Spatial Augmentation}
\author{Zidong Cao$^{1}$ \quad Jinjing Zhu$^{1}$ \quad Weiming Zhang$^{1}$ \quad Hao Ai$^{2}$ \quad Haotian Bai$^{1}$ \\
Hengshuang Zhao$^{3}$ \quad Lin Wang$^{4\dag}$ 
\vspace{1mm}\\
$^{1}$ AI Thrust, HKUST(GZ) \quad $^{2}$University of Birmingham \quad $^{3}$HKU \quad $^{4}$NTU
\vspace{1mm}\\}

\begin{document}
\twocolumn[{%
\renewcommand\twocolumn[1][]{#1}%
\vspace{-12mm}
\maketitle
\vspace{-10mm}
\begin{center}
    \captionsetup{type=figure}
    \includegraphics[width=0.95\linewidth]{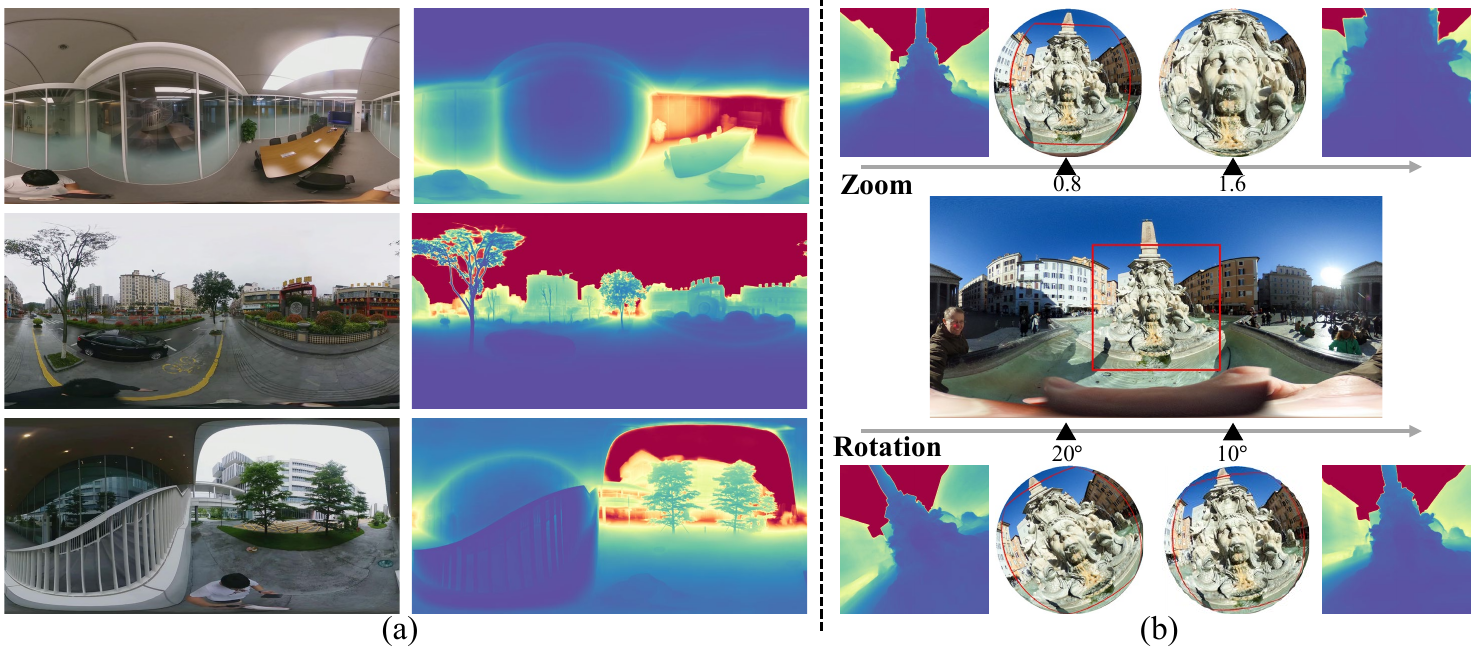}
    \vspace{-3mm}
    \captionof{figure}{\textbf{(a)} Our \textcolor{orange}{PanDA} exhibits impressive panoramic depth estimation results in open-world scenarios. The resolution of presented panoramas is 1008$\times$2016. \textbf{(b)} \textbf{Top row}: Spherical images with different zoom levels, and the corresponding depth predictions with perspective projection. \textbf{Middle row}: ERP image. \textbf{Bottom row}: Spherical images with different vertical rotation angles, and the corresponding depth predictions with perspective projection. Our \textcolor{orange}{PanDA} is robust to spherical transformations and predicts fine-grained depths.}\label{fig:teaser}
\end{center}%
}]
\begin{abstract}
\let\thefootnote\relax\footnote{$^{\dag}$ Corresponding author}Recently, Depth Anything Models (DAMs)~\cite{depthanything, yang2024depth} -- a type of depth foundation models --  have demonstrated impressive zero-shot capabilities across diverse perspective images. 
Despite its success, it remains an open question regarding DAMs' performance on panorama images that enjoy a large field-of-view ($180^{\circ}\times 360^{\circ}$) but suffer from spherical distortions. 
To address this gap, 
we conduct an empirical analysis to evaluate the performance of DAMs on panoramic images and identify their limitations.
For this, we undertake comprehensive experiments to assess the performance of DAMs from three key factors: panoramic representations, 360$^{\circ}$ camera positions for capturing scenarios, and spherical spatial transformations. This way, we reveal some key findings, \eg, DAMs are sensitive to spatial transformations. We then propose a semi-supervised learning (SSL) framework to learn a panoramic DAM, dubbed \textbf{PanDA}. Under the umbrella of SSL, PanDA first learns a teacher model by fine-tuning DAM through joint training on synthetic indoor and outdoor panoramic datasets. 
Then, a student model is trained using large-scale unlabeled data, leveraging pseudo-labels generated by the teacher model.  
To enhance PanDA's generalization capability, 
M\"obius transformation-based spatial augmentation (\textbf{MTSA}) is proposed to impose consistency regularization between the predicted depth maps from the original and spatially transformed ones.
This subtly improves the student model's robustness to various spatial transformations, even under severe distortions.  
Extensive experiments demonstrate that PanDA exhibits remarkable zero-shot capability across diverse scenes, and outperforms the data-specific panoramic depth estimation methods on two popular real-world benchmarks. Project page: \url{https://caozidong.github.io/PanDA_Depth/}.
\end{abstract}    
\vspace{-5mm}
\section{Introduction}
\label{sec:intro}

 360$^{\circ}$ cameras have gained significant interest for their ability to capture surrounding environments in a single shot~\cite{ai2025survey}. Monocular panoramic depth estimation is a crucial task for 3D scene perception with various applications, such as virtual reality (VR)~\cite{feng2022360} and autonomous driving~\cite{schon2021mgnet}. However, compared with normal perspective images, acquiring large-scale accurate depth annotations is much more expensive and difficult. Therefore, previous panoramic depth datasets,~\eg,~\cite{chang2017matterport3d, armeni2017joint, albanis2021pano3d}, are always scene-specific, especially limited to indoor scenes such as rooms. 
This limitation poses a significant challenge for current panoramic depth estimation methods~\cite{Li2022OmniFusion3M, Ai2023HRDFuseM3, Jiang2021UniFuseUF} when applied to real-world outdoor scenes~\cite{rey2022360monodepth}.

\begin{figure}[t]
    \centering
\includegraphics[width=\linewidth]{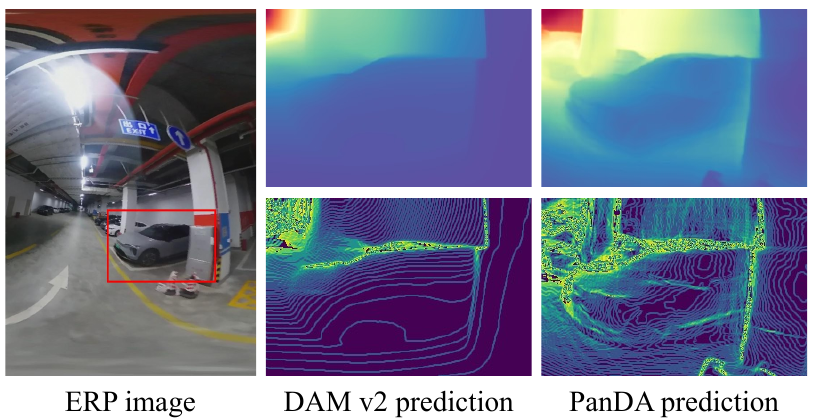}
\vspace{-7mm}
\caption{\textbf{Left}: Cropped patch of a panorama with 30$^{\circ}$ vertical rotation. \textbf{Top row}: Depth predictions. \textbf{Bottom row}: Gradient maps of depth predictions to better illustrate the depth variances. PanDA predicts clearer depth boundaries of the car.}
\vspace{-5mm}
\label{fig:Intro_figure}
\end{figure}

Recently, vision foundation models~\cite{kirillov2023segment,ke2024segment,depthanything} have been developed to address a wide range of vision tasks. Within the domain of monocular depth estimation, several foundational models have been introduced~\cite{ranftl2020towards, birkl2023midas, bhat2023ZoeDepth, depthanything, ke2024repurposing}. Among these, Depth Anything Models (DAMs)~\cite{depthanything, yang2024depth} stand out by leveraging large-scale unlabeled data, demonstrating high effectiveness. Despite the success of DAMs on perspective images, its performance on panoramas remains unclear. Panoramas are naturally different from perspective images, \ie, the large field-of-view (FoV) of $180^{\circ}\times 360^{\circ}$ and spherical distortions generated by sphere-to-plane projection~\cite{zioulis2018omnidepth}. This motivates us to investigate how DAMs perform when directly applied to panoramic images. To this end, we conduct an empirical investigation into several critical factors that influence DAMs' performance on panoramas: \textbf{1)} \textit{Different representations of panoramas}: The choice of representations is vital for the model to learn effective features. Panoramas can be represented in various representations, such as equirectangular projection (ERP), cubemap projection (CP), and tangent projection (TP), each offering distinct advantages and disadvantages in terms of FoVs and distortion levels. \textbf{2)} \textit{Different 360$^{\circ}$ camera positions:} For a given scene, varying camera heights and locations can alter the distance of objects from the camera and their corresponding latitudes in the spherical projection, significantly affecting the appearance of the captured panorama. \textbf{3)} \textit{Various spherical spatial transformations:} Given that panoramas support free viewing directions and immersive experiences~\cite{cao2023omnizoomer}, ensuring the robustness of depth estimation under sphere-based spatial transformations is crucial for real-world applications.

Our analysis reveals several key findings: \textbf{1)} The ERP representation outperforms other formats for DAMs, striking a balance between global consistency and local detail accuracy. \textbf{2)} Varying camera positions within the same scene can alter the layout of objects in the panorama, which may cause DAMs to fail, particularly when polar regions dominate the image. \textbf{3)} DAMs exhibit limited robustness to spatial transformations (See Fig.~\ref{fig:Intro_figure}). To address these challenges, we propose a semi-supervised learning (SSL) framework to develop a \textbf{pan}oramic \textbf{DA}M, termed \textbf{PanDA}. 
Within the SSL paradigm, PanDA first trains a teacher model by fine-tuning DAM v2~\cite{yang2024depth} with the Low-Rank Adaptation (LoRA)~\cite{hu2021lora, zhu2024melo}. To preserve fine-grained structural details and encompass diverse scenes, the teacher model is jointly trained on synthetic indoor and outdoor panorama datasets~\cite{zheng2020structured3d, li2022mode} to produce normalized depth outputs. Moreover, to enhance depth accuracy in content-rich equatorial regions, we introduce an equator-aware patch normalization loss (\textbf{EPNL}). EPNL focuses on sampling patches centered at the equator and performing local depth normalization, thereby decoupling normalization at the equator from the polar regions. This prevents essential structural details at the equator from being squeezed by global image normalization (See Tab.~\ref{tab:ablation-supervision_loss}). Subsequently, we train a student model using pseudo-labels generated by the teacher model. To harness the potential of large-scale unlabeled panoramas, we propose M\"obius transformation-based spatial augmentation (\textbf{MTSA}) to impose consistency regularization between the unlabeled data and spatially transformed ones. The MTSA enhances the student model’s robustness on spherical spatial transformation and improves its feature representations for objects affected by distortions (See Fig.~\ref{fig:Intro_figure} and Tab.~\ref{tab:ablation-ssl_loss}). Extensive experiments validate the effectiveness of PanDA across various spatial transformations and diverse scenes.

In summary, our contributions are three-fold: (\textbf{I}) We conduct a thorough analysis to evaluate the performance of DAMs on panoramas. (\textbf{II}) Informed by insights from our analysis, we develop an SSL framework named PanDA, which leverages large-scale unlabeled panoramas to enhance generalization. The proposed EPNL loss improves the depth accuracy at the equator, while MTSA increases the robustness of spatial transformations and enhances feature representations under distortions. (\textbf{III}) Experimental results show the impressive zero-shot capability of PanDA for being a panoramic depth foundation model, handling diverse scenes and various spatial transformations.

\section{Related Work}
\label{sec:related_work}

\noindent \textbf{Monocular Panoramic Depth Estimation.}
With the advance of deep learning and panoramic depth datasets~\cite{chang2017matterport3d,armeni2017joint,zioulis2018omnidepth}, monocular panoramic depth estimation methods have obtained good performance in specific datasets~\cite{chang2017matterport3d, armeni2017joint, zioulis2018omnidepth, albanis2021pano3d}. Previous methods mainly focus on mitigating the negative effects of distortion. For example, they have carefully designed distortion-aware convolution kernels~\cite{coors2018spherenet, Shen2022PanoFormerPT}, considered spherical prior~\cite{yun2023egformer}, or transformed the ERP image into distortion-less representations, \eg, cube map~\cite{Wang2020BiFuseM3} and tangent patches~\cite{Ai2023HRDFuseM3, Li2022OmniFusion3M}, and narrow FoV slices~\cite{Sun2020HoHoNet3I, pintore2021slicenet, yu2023panelnet}. However, as most panoramic depth datasets are captured in indoor scenes with limited data, these methods are difficult to generalize to unseen scenes, especially outdoor scenes~\cite{rey2022360monodepth}. Recently, Depth Anywhere~\cite{wang2024depth} utilizes large-scale unlabeled panoramas with pseudo labels from pre-trained DAM v1~\cite{depthanything} to improve the generalization capability~\cite{Jiang2021UniFuseUF,wang2022bifuse++}. \textit{Instead, we first investigate how DAMs perform on panoramas and
undertake comprehensive studies to evaluate DAMs. Then, in the SSL pipeline, we propose the EPNL to emphasize the equator and MTSA to improve
robustness.}

\begin{figure}[t]
    \centering
\includegraphics[width=\linewidth]{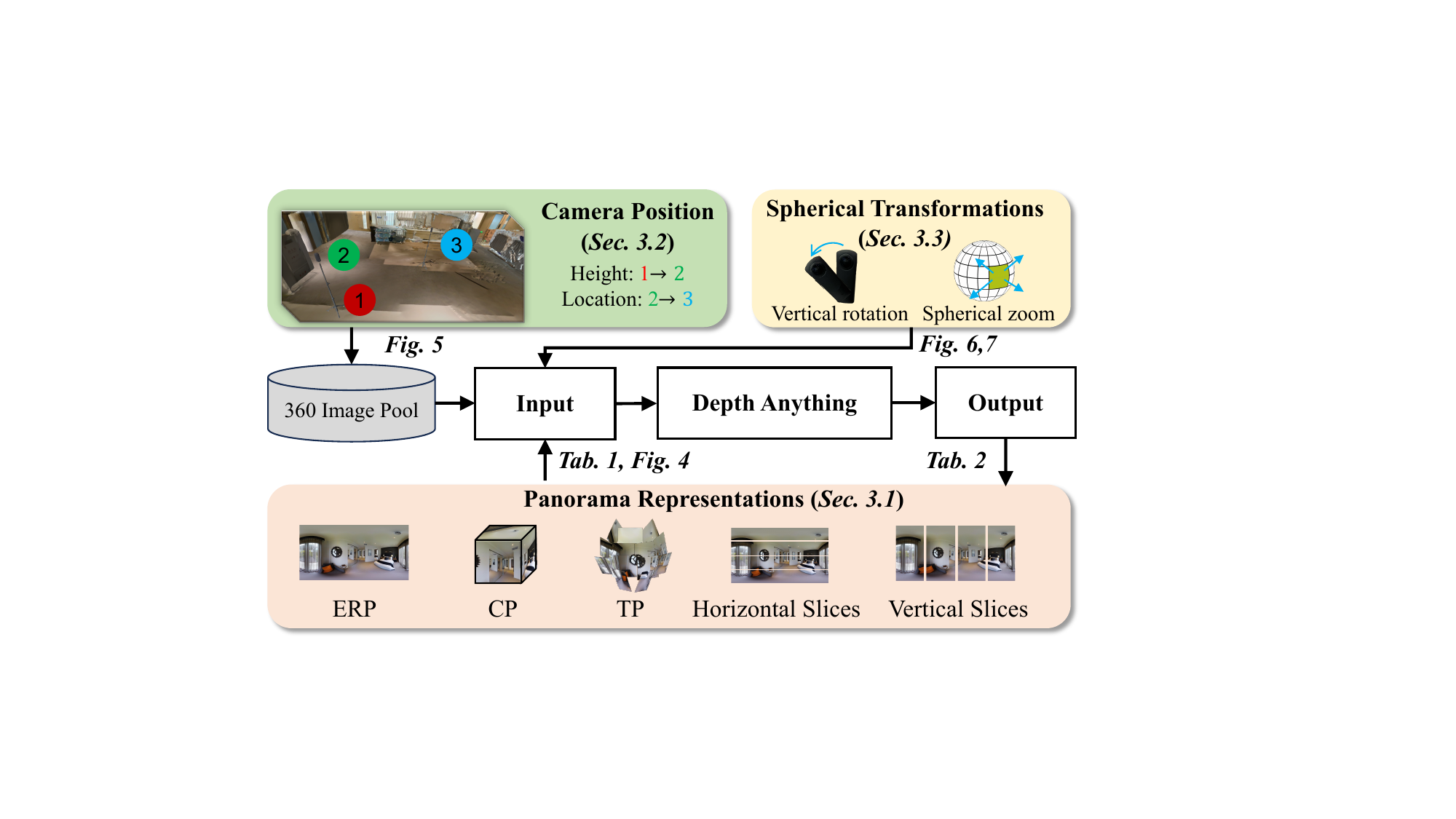}
\vspace{-7mm}
\caption{Overview of the analysis of DAMs.}
\vspace{-4mm}
\label{fig:benchmark_overview}
\end{figure}

\begin{table}[t]
    \centering
        \tablestyle{1.5pt}{1.05}
        \begin{tabular}{y{18mm}|x{12mm}|x{9mm}x{9mm}x{9mm}x{9mm}x{9mm}}
\toprule
Method & Backbone & ERP & CP & TP & HS & VS \\ 
\midrule
\multirow{3}{*}{DAM v1~\cite{depthanything}} & ViT-S & \cellcolor[HTML]{efefef}\textbf{0.1687} & 0.2144 & 0.2289 & 0.2104 & 0.1873\\
& ViT-B & \cellcolor[HTML]{efefef}\textbf{0.1629}  & 0.2238 & 0.2251 & 0.2073 & 0.1889\\
& ViT-L & \cellcolor[HTML]{efefef}\textbf{0.1614} & 0.2165 & 0.2046 & 0.2043 & 0.1858\\
\midrule
\multirow{3}{*}{DAM v2~\cite{yang2024depth}} & ViT-S & \cellcolor[HTML]{efefef}\textbf{0.1692} & 0.2205 & 0.2317 & 0.2186 & 0.1962 \\
& ViT-B & \cellcolor[HTML]{efefef}\textbf{0.1662} & 0.2249 & 0.2460 & 0.2149 & 0.2006 \\
& ViT-L & \cellcolor[HTML]{efefef}\textbf{0.1654} & 0.2238 & 0.2363 & 0.2101 & 0.1984 \\
\bottomrule
\end{tabular}
        \vspace{-2.5mm}
        \caption{Quantitative comparison of depth predictions from different panoramic representations after projecting to the ERP plane.}
        \label{tab:analysis_erp}
        \vspace{-3pt}
\end{table}

\noindent \textbf{Zero-shot Monocular Perspective Depth Estimation.}
To enhance the zero-shot capability of the monocular depth estimation model, MiDaS~\cite{ranftl2020towards, birkl2023midas} proposes to train on multiple perspective depth datasets. To mitigate the gap between different datasets, it introduces an affine-invariant loss to decouple depth scale and thereby focuses on the distribution consistency between the depth prediction and ground truth. Following this direction, ZoeDepth~\cite{bhat2023ZoeDepth} combines disparity and metric depth estimation together. ZoeDepth first trains a disparity depth estimation model on several datasets, and then fine-tunes it to metric depth estimation. Recently, Depth Anything v1~\cite{depthanything} and v2~\cite{yang2024depth} leverage large-scale unlabeled perspective images to enhance the representation capability of the model with semi-supervised learning. There are also depth estimation methods~\cite{ke2024repurposing} that utilize the rich knowledge of the visual world contained in Stable Diffusion~\cite{rombach2022high}. \textit{Our PanDA fully utilizes large-scale unlabeled data similar to Depth Anything.}

\begin{figure}[t]
    \centering
\includegraphics[width=\linewidth]{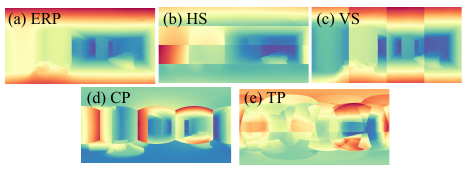}
\vspace{-7mm}
\caption{Different panoramic representations and their predicted depths after projecting back to the ERP plane.}
\vspace{-3mm}
\label{fig:Representation}
\end{figure}

\begin{table}[t]
    \centering
        \tablestyle{1.5pt}{1.05}
        \begin{tabular}{y{24mm}|x{18mm}x{18mm}x{18mm}}
\toprule
Inp. $\rightarrow$ Out.  &
Equator& Pole & Average \\ 
\midrule
ERP $\rightarrow$ CP & \cellcolor[HTML]{efefef}\textbf{0.1129} & \cellcolor[HTML]{efefef}\textbf{0.1201} & \cellcolor[HTML]{efefef}\textbf{0.1153} \\
CP \; $\rightarrow$ CP & 0.1164 & 0.1357& 0.1228 \\
\midrule
ERP $\rightarrow$ TP & \cellcolor[HTML]{efefef}\textbf{0.1235} & \cellcolor[HTML]{efefef}\textbf{0.1232} & \cellcolor[HTML]{efefef}\textbf{0.1234} \\
TP \; $\rightarrow$ TP & 0.1416 & 0.1492 & 0.1441 \\
\midrule
ERP $\rightarrow$ HS & \cellcolor[HTML]{efefef}\textbf{0.1322}  & \cellcolor[HTML]{efefef}\textbf{0.0965} & \cellcolor[HTML]{efefef}\textbf{0.1145} \\
HS \; $\rightarrow$ HS  & 0.1760 & 0.1251 & 0.1507 \\
\midrule
ERP $\rightarrow$ VS & \multirow{2}{*}{$-$} & \multirow{2}{*}{$-$} & 0.1438 \\
VS \; $\rightarrow$ VS & & &  \cellcolor[HTML]{efefef}\textbf{0.1355} \\
\bottomrule
\end{tabular}
        \vspace{-2.5mm}
        \caption{Quantitative comparison of depth predictions from different panoramic representations on different output spaces. We report the performance of all patches on average, and patches located at the equator and polar regions, respectively.}
        \vspace{-7pt}
        \label{tab:analysis_representation}
\end{table}

\noindent \textbf{Semi-supervised Learning (SSL).} SSL~\cite{van2020survey, yang2022survey} aims to leverage a large amount of unlabeled data to improve learning performance with limited labeled samples. Consequently, SSL has been applied to various tasks over the past decade, including image classification~\cite{wang2022usb, zhu2023patch}, object detection~\cite{sohn2020simple, xu2021end}, semantic segmentation~\cite{rangnekar2023semantic, ouali2020semi}, and depth estimation~\cite{baek2022semi, kuznietsov2017semi}. Inspired by the success of SSL in these tasks, our work aims to leverage large-scale unlabeled panoramas. \textit{Under the umbrella of SSL, we propose the MTSA to impose consistency regularization between the unlabeled panoramas and spatially transformed ones.}

\section{Analysis of Depth Anything Model}
\label{sec:analysis}

\textbf{Overview.} As illustrated in Fig.~\ref{fig:benchmark_overview}, to comprehensively assess the performance of DAMs on panoramas and identify their limitations, we conduct extensive experiments, analyzing several critical factors: (i) different panoramic representations, (ii) varying camera positions for capturing scenarios, and (iii) a range of spherical spatial transformations. Finally, we distill the key factors that hinder the effectiveness of DAM on panoramas, providing a theoretical basis for our PanDA model.

\noindent \textbf{Protocol.} We evaluate DAMs on the popular dataset, Matterport3D~\cite{chang2017matterport3d} as well as across various real-world scenarios. For the quantitative analysis, we report the Root Mean Squared Error (\textit{RMSE}) as the evaluation metric.

\subsection{Different Panoramic Representations}
\label{subsec:representation}

The choice of panoramic representations is crucial for the model performance~\cite{Jiang2021UniFuseUF, Ai2023HRDFuseM3, Li2022OmniFusion3M}. In this analysis, we select the most commonly used planar representation for panoramas--equirectangular projection (ERP), alongside four other distortion-free representations: cubemap patches (CP)~\cite{Jiang2021UniFuseUF}, tangent patches (TP)~\cite{Li2022OmniFusion3M}, horizontal slices (HS), and vertical slices (VS)~\cite{pintore2021slicenet} (See Fig.~\ref{fig:benchmark_overview}). Notably, all outputs are converted to the ERP plane for evaluation. As illustrated in Tab.~\ref{tab:analysis_erp}, although the ERP input contains greater distortion compared to other representations, it consistently achieves the best performance across different backbones. This superior performance can be attributed to the ability of ERP to maintain continuous and complete semantic content. Meanwhile, as illustrated in Fig.~\ref{fig:Representation}, depth predictions from patch-based or slice-based inputs exhibit significant discrepancies, leading to overall performance degradation. 

\begin{figure}[t]
    \centering
\includegraphics[width=\linewidth]{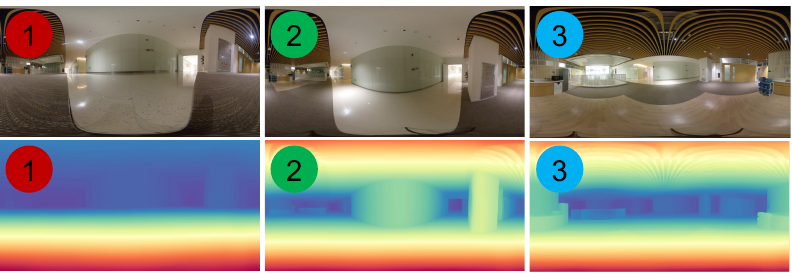}
\vspace{-7mm}
\caption{We place the 360$^{\circ}$ camera at three different heights and locations. (\textcolor{red}{\textbf{1}}) Positioning on the ground. (\textcolor{ForestGreen}{2}) Placing on the tripod. (\textcolor{blue}{\textbf{3}}) Magnifying towards the desk. Results show that the occupancy of polar regions influences the depth estimation at the equator. }
\vspace{-5mm}
\label{fig:error_height}
\end{figure}

To further assess the local details captured by outputs from different representations, we convert the ERP output into alternative formats and evaluate performance across distinct regions. In Tab.~\ref{tab:analysis_representation}, taking ERP as input achieves higher prediction accuracy compared to CP, TP and HS. However, when using ERP as the input, the accuracy in the VS output space is lower compared to using VS directly as the input. These results suggest that DAMs are effective at leveraging the available semantic content for depth estimation; the richer and more continuous the semantic content, the better the performance of DAMs.

\subsection{Different Camera Positions}
\label{subsec:camera_position}

In a given scenario, varying the 360$^{\circ}$ camera position can result in significant differences in the captured panoramas. As shown in the point cloud of Fig.~\ref{fig:benchmark_overview}, we conduct experiments using three distinct camera placements. Firstly, as shown in Fig.~\ref{fig:error_height}, positioning the camera directly on the ground causes the ground to dominate a large portion of the panorama, leading to poor depth predictions for objects located at the equator, such as desks. Next, elevating the camera using a tripod reduces the ground’s dominance in the polar region, significantly improving depth predictions for objects like desks and pillars. Finally, by placing the camera tripod closer to specific objects, such as the desk, these objects appear magnified in the panorama, allowing for more accurate depth predictions with finer structural details. Therefore, if the polar region dominates the panorama, the performance of DAM will be influenced, resulting in blurry structures at the equator.

\begin{figure}[t]
\centering
\includegraphics[width=\linewidth]{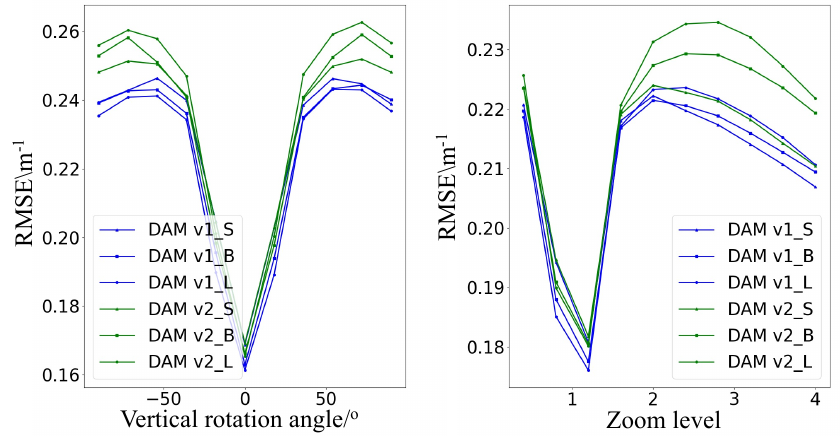}
\vspace{-7mm}
\caption{The performance of DAMs under different vertical rotation angles (\textbf{Left}) and various zoom level (\textbf{Right}).}
\label{fig:mobius_plot}
\end{figure}

\begin{figure}[t]
    \centering
\includegraphics[width=\linewidth]{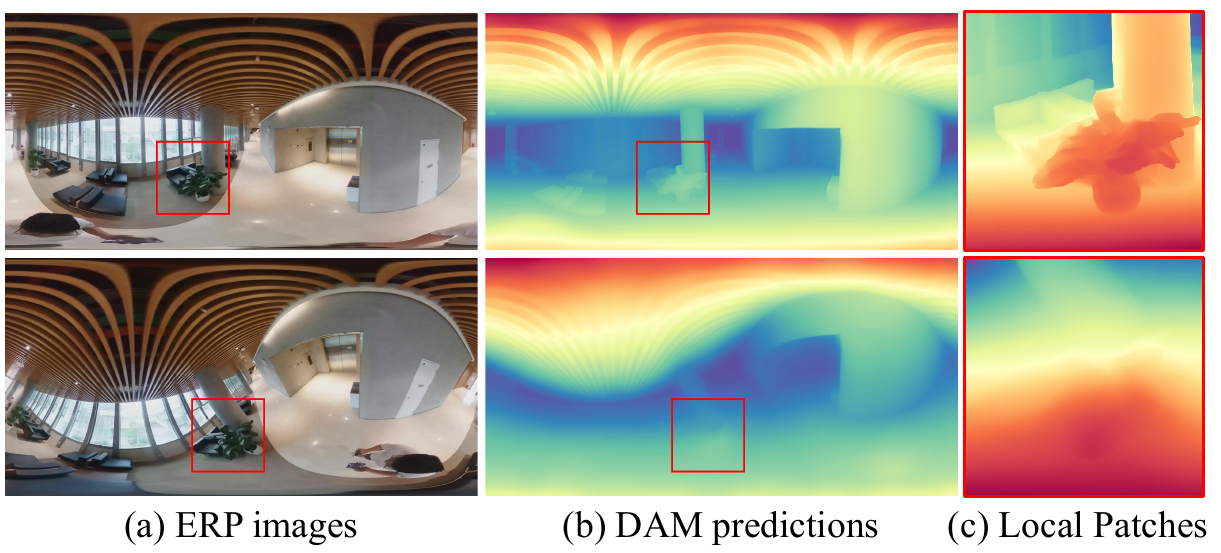}
\vspace{-7mm}
\caption{\textbf{Top row}: Original panorama and depth prediction. \textbf{Bottom row}: Panorama with 30$^{\circ}$ vertical rotation and depth prediction. The local patch is colored based on local normalization. The performance of DAM degrades after spatial transformation.}
\vspace{-5mm}
\label{fig:rotation_dam}
\end{figure}

\subsection{Various Spatial Transformations}
\label{subsec:spatial_transformation}

 As panoramas can offer free view direction and immersive experience, it is essential to evaluate the robustness of DAMs under various spherical spatial transformations.
In this analysis, we focus on two key transformations: vertical rotation and zoom (See Fig.~\ref{fig:benchmark_overview}). Note that horizontal rotation is equivalent to horizontal rolling. Specifically, we implement 11 vertical rotation angles with range from $-90^{\circ}$ to $90^{\circ}$, and 10 zoom levels ranging from $0.4$ to $4.0$. These transformations are achieved with the M\"obius transformation, which is the only conformal and bijective transformation on spherical surfaces~\cite{geyer2003conformal, schleimer2016squares, cao2023omnizoomer}. As shown in Fig.~\ref{fig:mobius_plot}, the performance of DAMs decreases significantly when subjected to spatial transformations. Specifically, as the absolute rotation angle increases, performance declines sharply, stabilizing with only minor variations beyond approximately $40^{\circ}$. Notably, the effects of upward and downward rotations are nearly symmetric. Regarding zoom transformations, both increasing and decreasing the zoom level adversely impact DAM performance. Interestingly, when the zoom level exceeds 2.4, higher zoom operation can even enhance performance. It could be attributed to the enlarged regions dominating the scene, thereby simplifying its overall structure and reducing the difficulty of depth estimation. In Fig.~\ref{fig:rotation_dam}, it can be found that the depth prediction of DAM degrades significantly.

\noindent \textbf{Summary}: Our analysis reveals findings: \textbf{1)} ERP representation outperforms other formats with a balance of global consistency and local detail preserving. \textbf{2)} If the polar region dominates the panorama, the depth prediction of DAM will be greatly influenced. \textbf{3)} The robustness of DAMs for various spatial transformations has to be improved.

\begin{figure}[t]
    \centering
\includegraphics[width=\linewidth]{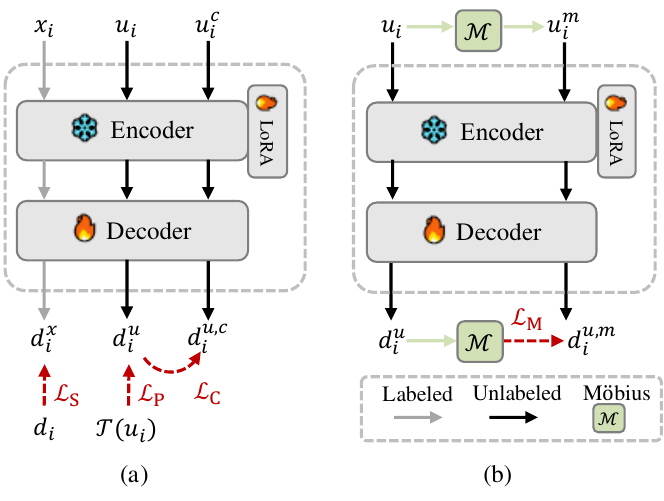}
\vspace{-7mm}
\caption{\textbf{(a)} The SSL pipeline on joint labeled and unlabeled datasets. \textbf{(b)} The M\"obius transformation-based spatial augmentation (MTSA) pipeline on unlabeled datasets. Our student model training combines \textbf{(a)} and \textbf{(b)} together.}
\vspace{-5mm}
\label{fig:method}
\end{figure}

\section{Methodology}
\label{sec:Model_Design}

\noindent \textbf{Overview.} Based on the findings in Sec.~\ref{sec:analysis}, we aim to develop a panoramic DAM by semi-supervised learning. We leverage labeled panoramic sets and large-scale unlabeled panoramic sets. Formally, 
we denote the labeled sets as $\mathcal{D}^l=$ $\left\{\left(x_i, d_i\right)\right\}_{i=1}^M$. We first train a teacher model $\mathcal{T}$ by fine-tuning DAM on $\mathcal{D}^l$ to produce pseudo depth labels for unlabeled sets. The unlabeled sets with pseudo labels can be denoted as $\mathcal{D}^u=\left\{(u_i,\mathcal{T}(u_i))\right\}_{i=1}^N$. $M$ and $N$ are numbers of samples.
After that, we train a student model $\mathcal{S}$ on the combination of $\mathcal{D}^l$ and $\mathcal{D}^u$.
Next, we introduce the fine-tuning on labeled dataset.

\subsection{Fine-tuning DAM with Labeled Data}
\label{sec: Fine-tuning}

As depicted in Tab.~\ref{tab:dataset}, labeled panoramic depth sets are considerably less than perspective datasets. To maintain the robust zero-shot capability of DAM while adapting it to panoramas, we fine-tune its encoder using Low-Rank Adaptation (LoRA)~\cite{hu2021lora,zhu2024melo}. 
We are inspired from~\cite{yang2024depth} that 
training on synthetic data can retain fine-grained structural details, while noisy labels in real-world depth datasets would result in blurry depth boundaries. Therefore, we train our teacher model $\mathcal{T}$ using two synthetic datasets: the Structured3D dataset~\cite{zheng2020structured3d} for indoor scenes, and the Deep360 dataset~\cite{li2022mode} for outdoor scenes. To facilitate joint training across both datasets, we perform depth normalization~\cite{ke2024repurposing}:

\begin{equation}
    \hat{d} = \frac{d - d_2}{d_{98} - d_2},
    \label{eq:normalizaiton}
\end{equation}

\begin{table}[t]
    \begin{minipage}{\linewidth}
    \centering
        \tablestyle{1.5pt}{1.05}
        \begin{tabular}{y{27mm}x{12mm}x{12mm}x{12mm}x{12mm}}
\toprule
Datasets & Indoor & Outdoor & Label & Samples \\ 
\midrule
\multicolumn{5}{c}{\textit{Synthetic Datasets}} \\
\midrule
Structured3D~\cite{zheng2020structured3d} & \ding{51} & & \ding{51} & 18298 \\
Deep360~\cite{li2022mode} & & \ding{51} & \ding{51} & 2100 \\
\midrule
\multicolumn{5}{c}{\textit{Unlabeled Real-world Datasets}} \\
\midrule
ZInD~\cite{cruz2021zillow} & \ding{51} & & \ding{55} & 54034 \\
360+x~\cite{chen2024360+} & \ding{51} & \ding{51} & \ding{55} & 47956 \\
\bottomrule
\end{tabular}
        \vspace{-2.5mm}
        \caption{The utilized labeled synthetic datasets and unlabeled real-world datasets for teacher model and student model training.}\label{tab:dataset}
        \vspace{-3mm}
    \end{minipage}
\end{table}

\noindent where $d_2$ and $d_{98}$ are the 2\% and 98\% percentiles of the valid depth values. After the normalization, we clip the depth range into $[0.01, 1]$ for stable training.
In this case, our teacher model predicts affine-invariant depth maps.

\begin{table*}[t]
    \begin{minipage}{\textwidth}
    \centering
        \tablestyle{1.5pt}{1.05}
        \begin{tabular}{y{25mm}|x{14mm}|x{12mm}|x{12mm}x{12mm}|x{12mm}x{12mm}|x{12mm}|x{12mm}x{12mm}|x{12mm}x{12mm}}
\toprule
\multicolumn{2}{c|}{Datasets} & \multicolumn{5}{c|}{Matterport3D~\cite{chang2017matterport3d}} & \multicolumn{5}{c}{Stanford2D3D~\cite{armeni2017joint}} \\
\midrule
\multirow{2}{*}{Methods} & \multirow{2}{*}{Backbone} & \multirow{2}{*}{Original} & \multicolumn{2}{c|}{Vertical Angle $\theta$} & \multicolumn{2}{c|}{Zoom Level $s$} & \multirow{2}{*}{Original} & \multicolumn{2}{c|}{Vertical Angle $\theta$} & \multicolumn{2}{c}{Zoom Level $s$}\\
& & & 10$^\circ$ & 20$^\circ$ & 2.0 & 3.0 & & 10$^\circ$ & 20$^\circ$ & 2.0 & 3.0 \\ 
\midrule
Marigold~\cite{ke2024repurposing} & SD 2.0~\cite{rombach2022high} & 0.5745 & 0.6105 & 0.7079 & 0.8486 & 0.9594 & 0.5069 & 0.5486 & 0.6273 & 0.7336 & 0.8624 \\
\midrule
DAM v2~\cite{yang2024depth} & \multirow{3}{*}{ViT-S} & 0.6063 & 0.6548 & 0.7691 & 0.9028 & 1.0433 & 0.5041 & 0.5224 & 0.6346 & 0.7587 & 0.8922 \\
PanDA (Ours) & & \textbf{0.4915} & \textbf{0.5188} & \textbf{0.5706} & \textbf{0.6242} & \textbf{0.7461} & \textbf{0.3462} & \textbf{0.3915} & \textbf{0.4392} & \textbf{0.5145} & \textbf{0.6322} \\
\cellcolor[HTML]{efefef}$\Delta$ & & \cellcolor[HTML]{efefef}18.93\% & \cellcolor[HTML]{efefef}20.77\% & \cellcolor[HTML]{efefef}25.81\% & \cellcolor[HTML]{efefef}30.86\% & \cellcolor[HTML]{efefef}28.49\% & \cellcolor[HTML]{efefef}31.32\% & \cellcolor[HTML]{efefef}25.06\% & \cellcolor[HTML]{efefef}30.79\% & \cellcolor[HTML]{efefef}32.19\% & \cellcolor[HTML]{efefef}29.14\% \\
\midrule
DAM v2~\cite{yang2024depth} & \multirow{3}{*}{ViT-B} & 0.5665 & 0.6386 & 0.7919 & 0.9210 & 1.0330 & 0.4870 & 0.5387 & 0.7086 & 0.8107 & 0.9146 \\
PanDA (Ours) & & 0.\textbf{4855} & \textbf{0.5004} & \textbf{0.5233} & \textbf{0.5497} & \textbf{0.7187} & \textbf{0.3253} & \textbf{0.3401} & \textbf{0.3754} & \textbf{0.4435} & \textbf{0.6112} \\
\cellcolor[HTML]{efefef}$\Delta$ & & \cellcolor[HTML]{efefef}14.30\% & \cellcolor[HTML]{efefef}21.64\% & \cellcolor[HTML]{efefef}33.92\% & \cellcolor[HTML]{efefef}40.31\% & \cellcolor[HTML]{efefef}30.42\% & \cellcolor[HTML]{efefef}33.20\% & \cellcolor[HTML]{efefef}36.87\% & \cellcolor[HTML]{efefef}47.02\% & \cellcolor[HTML]{efefef}45.29\% & \cellcolor[HTML]{efefef}33.17\% \\
\midrule
DAM v1~\cite{depthanything} & \multirow{4}{*}{ViT-L} & 1.1431 & 1.1735 & 1.1870 & 1.1068 & 1.0832 & 0.7597 & 0.7977 & 0.8653 & 0.8616 & 0.9152 \\
DAM v2~\cite{yang2024depth} &  & 0.5522 & 0.6900 & 0.9386 & 0.9484 & 1.0710 & 0.4884 & 0.5600 & 0.7749 & 0.8122 & 0.9405 \\
PanDA (Ours) & & \textbf{0.4690} & \textbf{0.4818} & \textbf{0.4963} & \textbf{0.5136} & \textbf{0.6791} & \textbf{0.3260} & \textbf{0.3240} & \textbf{0.3403} & \textbf{0.4202} & \textbf{0.5878} \\
$\Delta$ & & \cellcolor[HTML]{efefef}15.07\% & \cellcolor[HTML]{efefef}30.17\% & \cellcolor[HTML]{efefef}47.13\% & \cellcolor[HTML]{efefef}45.85\% & \cellcolor[HTML]{efefef}36.60\% & \cellcolor[HTML]{efefef}33.25\% & \cellcolor[HTML]{efefef}42.14\% & \cellcolor[HTML]{efefef}56.08\% & \cellcolor[HTML]{efefef}48.26\% & \cellcolor[HTML]{efefef}37.50\% \\
\bottomrule
\end{tabular}
        \vspace{-2.5mm}
        \caption{\textbf{Zero-shot metric depth estimation}. For Marigold, we utilize its LCM version with a single denoising step. The datasets for evaluation mainly contain indoor scenarios. In this case, for DAM v1, we utilize its version that is fine-tuned on the NYU dataset~\cite{silberman2012indoor}. For DAM v2, we utilize its version that is fine-tuned on the Hypersim dataset~\cite{roberts2021hypersim}.  We also evaluate under different spatial transformations.}\label{tab:comparison}
        \vspace{-5mm}
    \end{minipage}
\end{table*}

Furthermore, to enhance the depth accuracy at the equator, we propose an equator-aware patch normalization loss (EPNL). As the depth normalization in~\eqref{eq:normalizaiton} is applied across the whole panorama, fine-grained details could be squeezed by the stretched polar regions (See Sec.~\ref{subsec:camera_position}). In this case, we randomly crop $K$ patches from prediction $d_{i}^{x}$ and ground truth depth $d_i$, following~\cite{yin2023metric3d}. Then, we perform depth normalization~\cite{ranftl2020towards} within each patch, denoted as $\mathcal{N}_j(\cdot), j=[1,K]$. The EPNL can be calculated as:

\begin{equation}
\begin{aligned}
    \mathcal{L}_{\text{EPNL}}(d_{i}^{x}, d_{i}) = \frac{1}{M} \frac{1}{K}  \sum_{i=1}^{M} \sum_{j=1}^{K} \left|\mathcal{N}_j(d_{i}^{x}) - \mathcal{N}_j(d_i)\right|.
\end{aligned}
\end{equation}

\begin{figure}[t]
    \centering
\includegraphics[width=\linewidth]{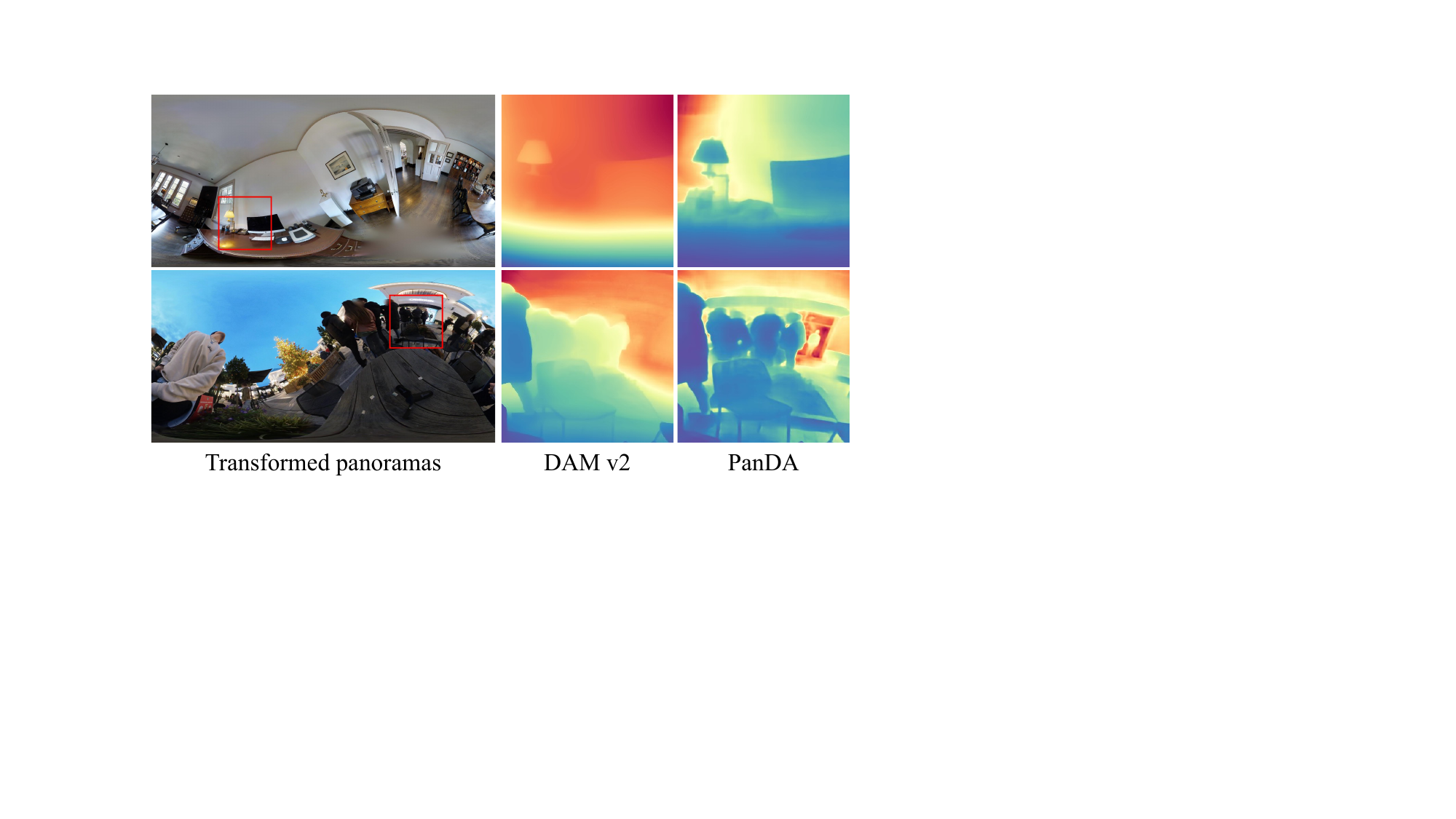}
\vspace{-7mm}
\caption{Qualitative comparison between DAM v2 and our PanDA for predictions of transformed panoramas.}
\label{fig:rotation_comp}
\end{figure}

\begin{figure*}[t]
    \centering
\includegraphics[width=\textwidth]{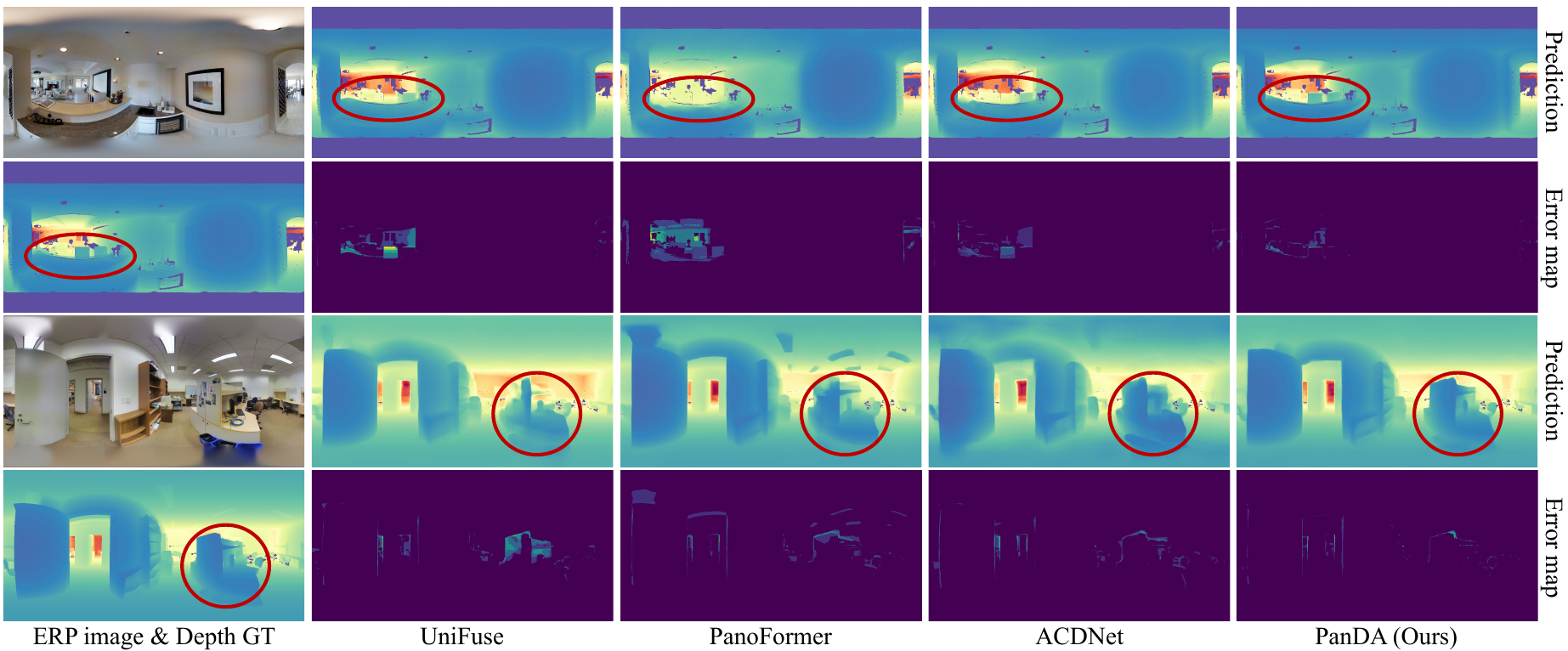}
\vspace{-7mm}
\caption{Qualitative comparison between SOTA monocular panoramic depth estimation methods and ours. We also provide error maps to illustrate the accuracy of structures of different methods. The top sample is from Matterport3D while the bottom one is from Stanford2D3D.}
\vspace{-5mm}
\label{fig:360_compare}
\end{figure*}

In Sec.~\ref{subsec:camera_position}, we find that polar regions could affect the depth performance at the equator. In addition, the equator region has the least distortion and enjoys rich semantic content. To facilitate depth prediction at the equator region, we propose sampling patches using a Gaussian distribution centered at the equator, rather than random sampling~\cite{yin2023metric3d}. Moreover, we ensure that the sampled patches are continuous at the left and right boundaries, preserving spherical coherence. In addition to EPNL, we also employ other supervised losses, such as SILog loss~\cite{eigen2014depth} $\mathcal{L}_\text{SILog}$ and gradient loss $\mathcal{L}_\text{Grad}$~\cite{yin2023metric3d}. The supervised loss can be formulated as:
\vspace{-2pt}
\begin{equation}
\begin{aligned}
    \label{eq:loss_s}
    \mathcal{L}_{\text{S}}(d_{i}^{x}, d_{i}) &= \mathcal{L}_\text{SILog}(d_{i}^{x}, d_{i})
    + \mathcal{L}_\text{Grad}(d_{i}^{x}, d_{i}) \\
    &+ \lambda_\text{E}\mathcal{L}_\text{EPNL}(d_{i}^{x}, d_{i}).
\end{aligned}
\end{equation}

\subsection{Semi-supervised Learning with Unlabeled Data}
\label{sec:ssl_learning}

After obtaining teacher model $\mathcal{T}$, we leverage it to  predict pseudo depth labels for unlabeled panoramas. In practice, the input resolution of unlabeled data is increased two times to improve the structural details in pseudo depths. We also employ SegFormer~\cite{xie2021segformer} to detect the \textit{sky} region, whose depth values are set to 1.0 (the farthest value of the normalized depth).

We then train a student model $\mathcal{S}$ on the combination of $\mathcal{D}^l$ and $\mathcal{D}^u$. As depicted in Fig.~\ref{fig:method}(a), 
we basically enforce the consistency between predictions from the teacher and student models. The form of the pseudo labeling loss function is consistent with the supervised loss $\mathcal{L}_\text{S}$:

\begin{equation}
    \mathcal{L}_{\text{P}} = \mathcal{L}_{\text{S}}(d_{i}^{u}, \mathcal{T}(u_i)).
\end{equation}

Furthermore, to harness the potential of large-scale unlabeled panoramas, we employ strong augmentations and enforce the consistency of predicted depths from the original and augmented panoramas. We first apply the strong color augmentation to $u_i$ to obtain $u_{i}^{c}$. Color augmentation includes color jittering to randomly adjust the brightness, contrast, saturation, \etc. Given the input $u_{i}^{c}$, we can obtain the prediction $d_{i}^{u,c}$, and consistency loss is formulated as:
\vspace{-1mm}
\begin{equation}
    \mathcal{L}_{\text{C}} = \mathcal{L}_\text{SILog}(d_{i}^{u,c},d_{i}^{u}).
\end{equation}

To improve the robustness of the student model to spatial transformations,  we propose the M\"obius  transformation-based spatial augmentation (MTSA), as illustrated in Fig.~\ref{fig:method}(b). Specifically, the M\"obius transformation depends on two factors: vertical rotation angle $\theta$ and zoom level $s$. By default, we uniformly sample $\theta$ in $[-10^{\circ}, 10^{\circ}]$, and $s$ in $[1.0, 1.5]$ during training. As illustrated in Fig.~\ref{fig:method}(b), for an input unlabeled panorama $u_i$, we conduct the M\"obius transformation, denoted as $\mathcal{M}(\cdot)$, resulting in a transformed image $u_{i}^{m}$. Both $u_i$ and $u_{i}^{m}$ are then passed through the student model to generate depth predictions $d_{i}^{u}$ and $d_{i}^{u,m}$, respectively. Subsequently, we apply $\mathcal{M}(\cdot)$ with the same factors to the depth prediction $d_{i}^{u}$ and enforce consistency regularization as follows: 
\vspace{-1mm}
\begin{equation}
    \mathcal{L}_{\text{M}} = \mathcal{L}_\text{SILog}(d_{i}^{u,m},\mathcal{M}(d_{i}^{u})).
\end{equation}

Overall, the semi-supervised loss $\mathcal{L}_{\text{SSL}}$ is: 

\begin{equation}\mathcal{L}_{\text{SSL}}=\mathcal{L}_{\text{S}}+\mathcal{L}_{\text{P}}+\lambda_\text{C}\mathcal{L}_{\text{C}}+\lambda_\text{M}\mathcal{L}_{\text{M}}. 
\label{eq:overall_loss}
\end{equation}

\noindent \textbf{Discussion.} The transformed panoramas have more severe distortions due to increased curves. To challenge the model to learn feature representations on these transformed samples, MTSA could enable the student model to better handle spherical distortions. Therefore, MTSA also benefits the original panoramic depth estimation (See Tab.~\ref{tab:ablation-ssl_loss}). 
\section{Experiment}
\label{experiment}

\begin{table}[t]
    \begin{minipage}{\linewidth}
    \centering
        \tablestyle{1.5pt}{1.05}
        \begin{tabular}{y{26mm}|x{11mm}x{11mm}|x{9mm}x{9mm}x{9mm}}
\toprule
Methods & \textit{AbsRel} $\downarrow$ & \textit{RMSE} $\downarrow$ & $\delta_1$ $\uparrow$ & $\delta_2$ $\uparrow$ & $\delta_3$ $\uparrow$ \\
\midrule
BiFuse~\cite{Wang2020BiFuseM3} & 0.2048 & 0.6259 & 84.52 & 93.19 & 96.32 \\
UniFuse~\cite{Jiang2021UniFuseUF} & 0.1063 & 0.4941 & 88.97 & 96.23 & 98.31 \\
HoHoNet~\cite{Sun2020HoHoNet3I} & 0.1488 & 0.5138 & 87.86 & 95.19 & 97.71 \\
BiFuse++~\cite{wang2022bifuse++} & $-$ & 0.5190 & 87.90 & 95.17 & 97.72 \\
ACDNet~\cite{zhuang2022acdnet} & 0.1010 & 0.4629 & 90.00 & 96.78 & 98.76 \\
PanoFormer~\cite{Shen2022PanoFormerPT} & 0.0904 & 0.4470 & 88.16 & 96.61 & 98.78 \\
HRDFuse~\cite{Ai2023HRDFuseM3} & 0.0967 & 0.4433 & 91.62 & 96.69 & 98.44 \\
S2Net~\cite{li2023mathcal} & 0.0865 & 0.4052 & 92.64 & 97.68 & 99.11 \\
EGFormer~\cite{yun2023egformer} & 0.1473 & 0.6025 & 81.58 & 93.90 & 97.35 \\
Elite360D~\cite{ai2024elite360d} & 0.1115 & 0.4875 & 88.15 & 96.46 & 98.74 \\
Depth Anywhere~\cite{wang2024depth} & 0.0850 & $-$ & 91.70 & 97.60 & 99.10 \\
\midrule
PanDA-S & 0.0922 & 0.3950 & 92.26 & 98.30 & 99.47 \\
PanDA-B & 0.0792 & 0.3475 & 94.60 & 98.75 & 99.60 \\
PanDA-L & \cellcolor[HTML]{efefef}\textbf{0.0717} & \cellcolor[HTML]{efefef}\textbf{0.3305} & \cellcolor[HTML]{efefef}\textbf{95.09} & \cellcolor[HTML]{efefef}\textbf{98.94} & \cellcolor[HTML]{efefef}\textbf{99.65} \\
\bottomrule
\end{tabular}
        \vspace{-2.5mm}
        \caption{Quantitative comparison on the Matterport3D dataset.}\label{tab:comparison_matterport3d}
        \vspace{-5mm}
    \end{minipage} \\
\end{table}

\subsection{Implementation Details}
\label{subsec:implementation}

\noindent \textbf{Datasets.} Datasets are summarized in Tab.~\ref{tab:dataset}. We leverage two real-world datasets--Matterport3D~\cite{chang2017matterport3d} and Stanford2D3D~\cite{armeni2017joint}--to access the zero-shot performance of PanDA in comparison with zero-shot depth foundation models. We also benchmark PanDA against SOTA panoramic depth estimation methods by fine-tuning it on real-world datasets. For training, we combine the datasets directly, repeating the labeled datasets to ensure the number of samples matches that of the unlabeled datasets. The training resolution is $504\times1008$. For a fair comparison, in Tab.~\ref{tab:comparison_matterport3d},~\ref{tab:comparison_stanford2d3d}, we resize our depth predictions to $512\times1024$.

\begin{table}[t]
    \begin{minipage}{\linewidth}
    \centering
        \tablestyle{1.5pt}{1.05}
        \begin{tabular}{y{26mm}|x{11mm}x{11mm}|x{9mm}x{9mm}x{9mm}}
\toprule
Methods & \textit{AbsRel} $\downarrow$ & \textit{RMSE} $\downarrow$ & $\delta_1$ $\uparrow$ & $\delta_2$ $\uparrow$ & $\delta_3$ $\uparrow$ \\
\midrule
BiFuse~\cite{Wang2020BiFuseM3} & 0.1209 & 0.4142 & 86.60 & 95.80 & 98.60 \\
UniFuse~\cite{Jiang2021UniFuseUF} & 0.1114 & 0.3691 & 87.11 & 96.64 & 98.82 \\
HoHoNet~\cite{Sun2020HoHoNet3I} & 0.1014 & 0.3834 & 90.54 & 96.93 & 98.86 \\
BiFuse++~\cite{wang2022bifuse++} & $-$ & 0.3720 & 87.83 & 96.49 & 98.84 \\
ACDNet~\cite{zhuang2022acdnet} & 0.0984 & 0.3410 & 88.72 & 97.04 & 98.95 \\
PanoFormer~\cite{Shen2022PanoFormerPT} & 0.1131 & 0.3557 & 88.08 & 96.23 & 98.55 \\
HRDFuse~\cite{Ai2023HRDFuseM3} & 0.0935 & 0.3106 & 91.40 & 97.98 & 99.27 \\
S2Net~\cite{li2023mathcal} & 0.0903 & 0.3383 & 91.91 & 97.82 & 99.12 \\
EGFormer~\cite{yun2023egformer} & 0.1528 & 0.4974 & 81.85 & 93.38 & 97.36 \\
Elite360D~\cite{ai2024elite360d} & 0.1182 & 0.3756 & 88.72 & 96.84 & 98.92 \\
Depth Anywhere~\cite{wang2024depth} & 0.1180 & 0.3510 & 91.00 & 97.10 & 98.70 \\
\midrule
PanDA-S & 0.0762 & 0.2866 & 95.31 & 98.60 & 99.36 \\
PanDA-B & 0.0635 & 0.2682 & 95.84 & 98.95 & 99.51 \\
PanDA-L & \cellcolor[HTML]{efefef}\textbf{0.0609} & \cellcolor[HTML]{efefef}\textbf{0.2540} & \cellcolor[HTML]{efefef}\textbf{96.82} & \cellcolor[HTML]{efefef}\textbf{99.05} & \cellcolor[HTML]{efefef}\textbf{99.52} \\
\bottomrule
\end{tabular}
        \vspace{-2.5mm}
        \caption{Quantitative comparison on the Stanford2D3D dataset.}\label{tab:comparison_stanford2d3d}
        \vspace{-5mm}
    \end{minipage}
\end{table}

\noindent \textbf{Implementation Details.} All experiments are conducted on A800 GPUs. The learning rate is set to 1e-4 using the Adam optimizer~\cite{kingma2014adam}. The teacher model is trained for 20 epochs, 
while the student model is trained for 4 epochs.
The loss weight $\lambda_\text{E}$ in \cref{eq:loss_s} is set to 5.0,  $\lambda_\text{C}$ and $\lambda_\text{M}$ in \cref{eq:overall_loss} are set to 2.0 and 1.0, respectively. In addition, fine-tuning on the two real-world datasets is conducted with 30 epochs. The batch size is 4. Data augmentation includes color jittering, horizontal translation, and flipping following~\cite{Jiang2021UniFuseUF}.

\noindent \textbf{Metric.} Following~\cite{Jiang2021UniFuseUF}, we evaluate depth estimation performance with metrics including Absolute Relative Error (\textit{AbsRel}), Root Mean Squared Error (\textit{RMSE}), and three percentage metrics $\delta_i$, where $i \in \{1.25, 1.25^2, 1.25^3\}$.

\subsection{Qualitative and Quantitative Evaluation}

\noindent \textbf{Comparison with Zero-shot Methods.} As illustrated in Tab.~\ref{tab:comparison}, we compare with zero-shot depth estimation methods designed for perspective images, \eg, DAM v1~\cite{depthanything}, DAM v2~\cite{yang2024depth}, and Marigold~\cite{ke2024repurposing}. The results demonstrate that our PanDA outperforms the other methods across all metrics and datasets, highlighting its effective zero-shot capability for real-world panoramic depth estimation. Regarding performance under various transformations, PanDA exhibits significantly less performance degradation. We attribute this to the proposed MTSA, which enables the model to learn spherical transformations during training. The model is enhanced to handle severe distortions. Fig.~\ref{fig:rotation_comp} also verifies that our PanDA is robust enough to various spherical transformations. Instead, performance of DAM v2
drops obviously when encountering spherical transformations.

\noindent \textbf{Comparison with SOTA Panoramic Methods.} We fine-tune PanDA to Matterport3D and Stanford2D3D datasets. As shown in Tab.~\ref{tab:comparison_matterport3d} and Tab.~\ref{tab:comparison_stanford2d3d}, our PanDA with ViT-B and ViT-L as backbones surpasses previous methods across all metrics, while PanDA with ViT-S as the backbone outperforms previous methods in most metrics. For instance, in the Matterport3D dataset, PanDA with ViT-L as the backbone outperforms HRDFuse \textit{RMSE} metric from 0.4433 $\rightarrow$ 0.3305. We ascribe it to our proposed fine-tuning strategies that adapt DAM to panoramas effectively. As shown in Fig.~\ref{fig:360_compare}, our PanDA predicts clear structural details such as chairs and desks, which are blurry in other methods. The error maps also demonstrate that PanDA can predict more accurate depths compared to previous panoramic methods.

\begin{table}[t]
    \begin{minipage}{\linewidth}
    \centering
        \tablestyle{1.5pt}{1.05}
        \begin{tabular}{y{15mm}|x{12mm}|x{12mm}x{12mm}|x{12mm}x{12mm}}
\toprule
\multirow{2}{*}{Methods} & \multirow{2}{*}{Backbone} & \multicolumn{2}{c}{Matterport3D~\cite{chang2017matterport3d}} & \multicolumn{2}{c}{Stanford2D3D~\cite{armeni2017joint}} \\
& & \textit{AbsRel} $\downarrow$ & \textit{RMSE} $\downarrow$ & \textit{AbsRel} $\downarrow$ & \textit{RMSE} $\downarrow$ \\
\midrule
Baseline & \multirow{3}{*}{ViT-S} & 0.1264 & 0.5132 & \cellcolor[HTML]{efefef}\textbf{0.1010} & 0.3418\\
+RPNL~\cite{yin2023metric3d} & & 0.1303 & 0.5268 & 0.1220 & 0.3533 \\
+EPNL & & \cellcolor[HTML]{efefef}\textbf{0.1256} & \cellcolor[HTML]{efefef}\textbf{0.5062} & 0.1109 & \cellcolor[HTML]{efefef}\textbf{0.3401} \\
\midrule
Baseline & \multirow{3}{*}{ViT-L} & 0.1057 & 0.4615 & \cellcolor[HTML]{efefef}\textbf{0.1082} & 0.3442 \\
+RPNL~\cite{yin2023metric3d} & & 0.1061 & 0.4660 & 0.1104 & 0.3398 \\
+EPNL & & \cellcolor[HTML]{efefef}\textbf{0.1036} & \cellcolor[HTML]{efefef}\textbf{0.4539} & 0.1092 & \cellcolor[HTML]{efefef}\textbf{0.3314} \\
\bottomrule
\end{tabular}
        \vspace{-2.5mm}
        \caption{Ablation studies for the proposed \textbf{EPNL}.}\label{tab:ablation-supervision_loss}
        \vspace{-5mm}
    \end{minipage} \\
\end{table}

\subsection{Ablation Studies}

\noindent \textbf{Effectiveness of EPNL.} In Tab.~\ref{tab:ablation-supervision_loss}, we verify the effectiveness of our proposed EPNL under two backbones. We train on Structured3D~\cite{zheng2020structured3d} and Deep360~\cite{li2022mode} datasets and perform zero-shot depth estimation evaluation on Matterport3D~\cite{chang2017matterport3d} and Stanford2D3D~\cite{armeni2017joint} datasets. The baseline is set with SILog loss and gradient loss. The results demonstrate that our EPNL outperforms the baseline and previous RPNL in most metrics. It shows the efficacy of local normalization. Instead of random sampling in the whole panorama, our proposed sampling strategy is more suitable for panoramic depth estimation by improving the priority of equator regions. We also ensure the continuity of left and right boundaries for ERP images for spherical coherence.

\noindent \textbf{Effectiveness of SSL Losses.} In Tab.~\ref{tab:ablation-ssl_loss}, we discuss the impact of SSL loss functions on student training. By default, the supervised loss $\mathcal{L}_{\text{S}}$ is employed. Firstly, incorporating unlabeled data with pseudo depth labels yields performance improvement. Subsequently, adding the consistency regularization with MTSA further improves the performance in the original condition and conditions under various transformations. Moreover, it can be found that only adding color augmentation (CA) has no obvious benefit for student model training. Instead, combining CA and MTSA together facilitates the student model to learn robust representations, resulting in the best performance for all conditions.

\begin{table}[t]
    \begin{minipage}{\linewidth}
    \centering
        \tablestyle{1.5pt}{1.05}
        \begin{tabular}{y{27mm}|x{10mm}|x{9.5mm}x{9.5mm}|x{9.5mm}x{9.5mm}}
\toprule
\multirow{2}{*}{Methods} & \multirow{2}{*}{Original} & \multicolumn{2}{c|}{Vertical angle $\theta$} & \multicolumn{2}{c}{Zoom level $s$} \\
& & $10^{\circ}$ & $20^{\circ}$ & 2.0 & 3.0 \\
\midrule
$\mathcal{L}_{s}$ & 0.5109 & 0.5711 & 0.6804 & 0.8381 & 0.9793 \\
\midrule
$\mathcal{L}_{s}+\mathcal{L}_{p}$ & 0.4977 & 0.5588 & 0.6678 & 0.8356 & 0.9854 \\
$\mathcal{L}_{s}+\mathcal{L}_{p}+\mathcal{L}_{m}$ & 0.4950 & 0.5198 & 0.5781 & 0.6396 & 0.7640 \\
\midrule
$\mathcal{L}_{s}+\mathcal{L}_{p}+\mathcal{L}_{c}$ & 0.4984 & 0.5671 & 0.6895 & 0.8318 & 0.9819 \\
$\mathcal{L}_{s}+\mathcal{L}_{p}+\mathcal{L}_{c}+\mathcal{L}_{m}$ & \cellcolor[HTML]{efefef}\textbf{0.4915} & \cellcolor[HTML]{efefef}\textbf{0.5188} & \cellcolor[HTML]{efefef}\textbf{0.5706} & \cellcolor[HTML]{efefef}\textbf{0.6242} & \cellcolor[HTML]{efefef}\textbf{0.7461} \\
\bottomrule
\end{tabular}
        \vspace{-2.5mm}
        \caption{Ablation studies for \textbf{semi-supervised training losses}. We report \textit{RMSE} metric on Matterport3D dataset.}\label{tab:ablation-ssl_loss}
        \vspace{2mm}
    \end{minipage}
\end{table}

\begin{figure}[t]
    \centering
\includegraphics[width=\linewidth]{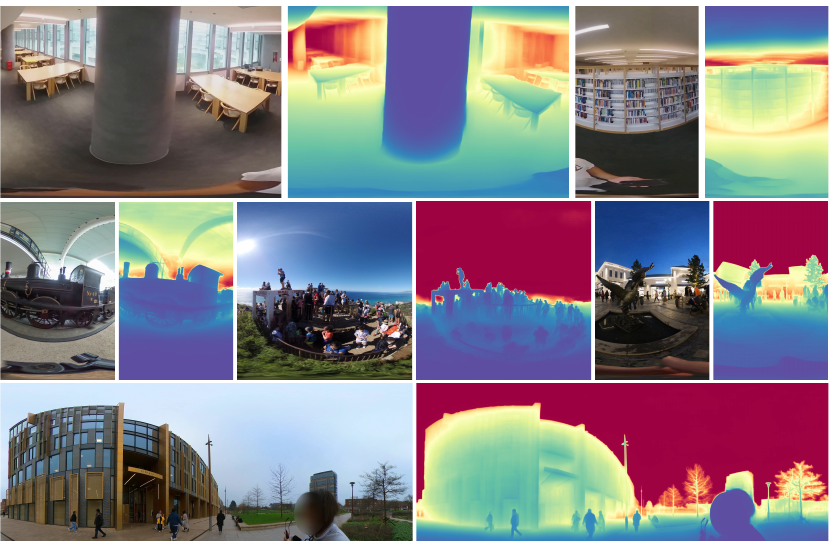}
\vspace{-7mm}
\caption{Qualitative results of PanDA on open-world scenes.}
\label{fig:real_world_samples}
\end{figure}

\subsection{Open-world Qualitative Results}
\label{subsec:experiment-openworld}

We provide more qualitative results for real-world samples in Fig.~\ref{fig:real_world_samples}. It can be observed that PanDA can predict clear depth boundaries in both indoor and outdoor scenes. 

\section{Conclusion}
\label{limitations}

\noindent In this paper, we provided a comprehensive analysis for evaluating the performance of DAMs on panoramas by exploring several key factors, \eg, panoramic image representations, 360$^{\circ}$ camera positions, and spherical spatial transformations. The analysis reveals some key findings, such as that DAMs are less robust to various spatial transformations. Leveraging these findings, we fine-tuned DAM and unleashed the potential of large-scale unlabeled panoramas under the umbrella of semi-supervised learning with M\"obius transformation-based spatial augmentation. The experiments demonstrate the impressive zero-shot capability of our method, establishing it as a potential depth foundational model for panoramas.

\noindent \textbf{Acknowledgement.} This paper is supported by the National Natural Science Foundation of China (NSF) under Grant No. 62206069 (affiliated with Guangzhou HKUST FYTRI) and the Start Up Grant at Nanyang Technological University (NTU) under Grant No. 03INS002165C140.

\clearpage
\clearpage
\setcounter{page}{9}
\maketitlesupplementary
\appendix

\section{Datasets}
\label{sec:datasets}

In this section, we explain the data processing of each dataset in detail.

\noindent \textbf{Structured3D dataset~\cite{zheng2020structured3d}.} It is an indoor synthetic dataset. We employ its training set, which consists of 18,298 samples, for our training purposes. In terms of data processing, we initially scale the depth map values by a factor of 0.001. Subsequently, we clip these values to a range of 0 to 10 meters. Finally, we apply depth normalization~\cite{ke2024repurposing}.

\noindent \textbf{Deep360 dataset~\cite{li2022mode}.} This dataset is synthetic and contains outdoor scenes, generated using the CARLA simulator~\cite{dosovitskiy2017carla}. It comprises pairs of fisheye images and depth maps. Following the official guidance of~\cite{li2022mode}, we transform the fisheye format into the ERP representation. We clip the depth values to a range of 0 to 100 meters. We restrict larger depth values in the sky region to 100 meters. Subsequently, depth normalization is applied following~\cite{ke2024repurposing}.

\noindent \textbf{ZInD dataset~\cite{cruz2021zillow}.} This is an indoor dataset with room layout annotations but lacking depth labels. We employ it for semi-supervised learning to enhance the scene diversity of indoor environments. We utilize its training set with 54034 samples for training.

\noindent \textbf{360+x dataset~\cite{chen2024360+}.} This dataset encompasses both indoor and outdoor scenes, showcasing its diversity. For data processing, we uniformly extract frames from the high-resolution videos contained within the 360+x dataset. From approximately 200 videos, we extract a total of 47,956 frames. Subsequently, we observe that the performance of our PanDA is suboptimal in extremely dark regions. As a result, these scenes are omitted from the training set. Finally, we utilize SegFormer~\cite{xie2021segformer} to detect sky regions and assign a depth value of 1.0 to these areas, which represents the maximum value on the normalized depth map.

\noindent \textbf{Matterport3D dataset~\cite{chang2017matterport3d}.} It is used to validate the effectiveness of our PanDA in real-world scenes. The maximum depth value is set at 10 meters.

\noindent \textbf{Stanford2D3D dataset~\cite{armeni2017joint}.} It is also used to validate the effectiveness of our PanDA in real-world scenes. The maximum depth is set at 10 meters. Given that the top and bottom parts of panoramas in the Stanford2D3D dataset are missing, we fill in these missing areas by following the methods described in UniFuse~\cite{Jiang2021UniFuseUF}. 

\noindent \textbf{Other datasets.} Besides Deep360~\cite{li2022mode}, there is another synthetic dataset~\cite{bhanushali2022omnihorizon} containing outdoor scenes. However, it was not publicly available at the time of submission.

  \begin{figure}[t]
    \centering
\includegraphics[width=0.8\linewidth]{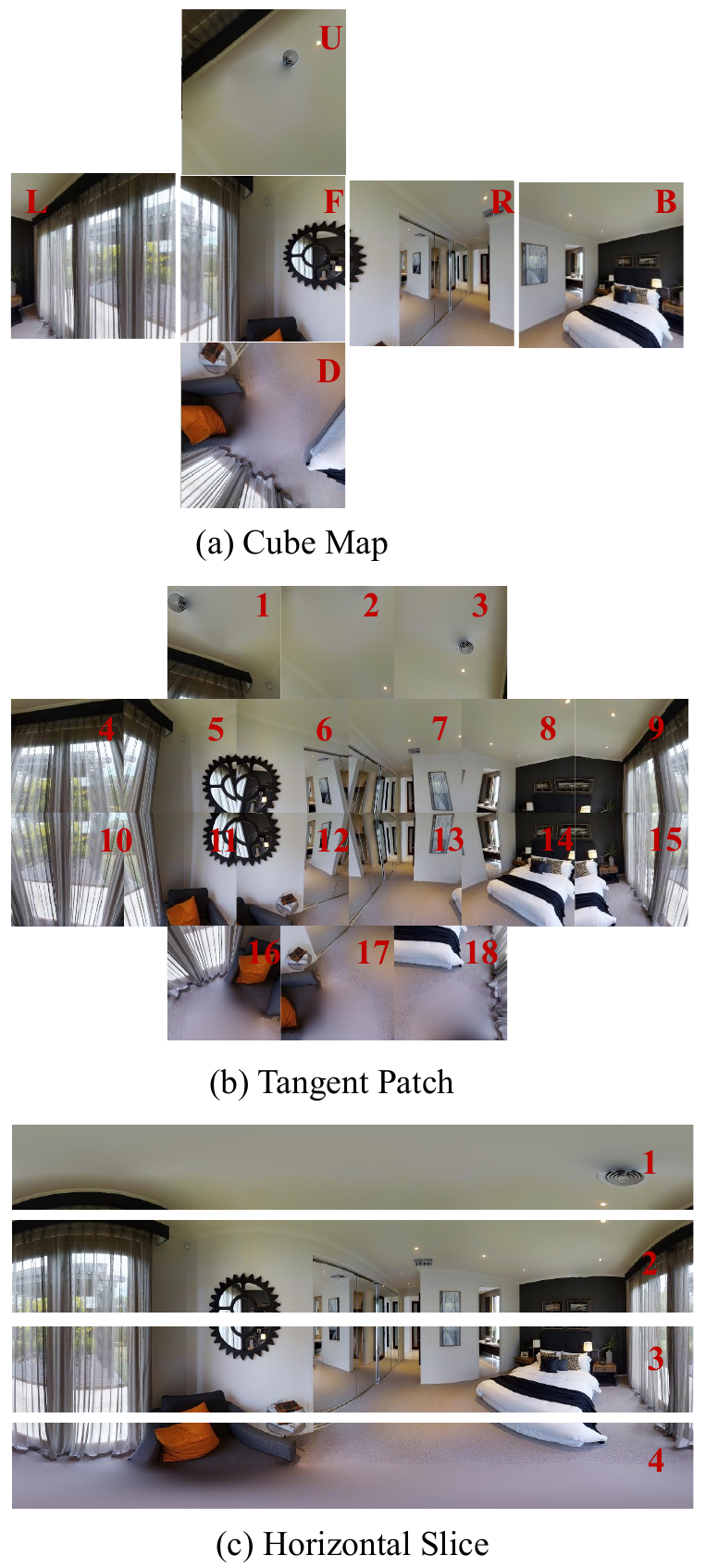}
\caption{Illustration of the indexes of patches in different panoramic representations.}
\label{fig:supp_index}
\end{figure}

\begin{table*}[t]
    \begin{minipage}{\textwidth}
    \centering
        \tablestyle{1.5pt}{1.05}
        \begin{tabular}{y{25mm}|x{20mm}|x{24mm}|x{24mm}|x{30mm}|x{40mm}}
\toprule
\textbf{Representation} & \textbf{Patch Number} & \textbf{FoV} & \textbf{Resolution} & Equator Region & Pole Region \\
\midrule
ERP & 1 & $180^{\circ} \times 360^{\circ}$ & $504 \times 1008$ & $-$ & $-$ \\
Cube map (CP) & 6 & $90^{\circ} \times 90^{\circ}$ & $252 \times 252$ & $\{$Front, Left, Right, Back$\}$ & $\{$Top, Down$\}$ \\
Tangent patch (TP) & 18 & $80^{\circ} \times 80^{\circ}$ & $126 \times 126$ & $4^{th}$ to $15^{th}$ & $1^{st}$ to $3^{rd}$, and $16^{th}$ to $18^{th}$\\
Horizontal slice (HS) & 4 & $45^{\circ} \times 360^{\circ}$ & $126 \times 1008$ & $2^{nd}$ and $3^{rd}$ & $1^{st}$ and $4^{th}$\\
Vertical slice (VS) & 4 & $180^{\circ} \times 90^{\circ}$ & $504 \times 252$ & $-$ & $-$\\
\bottomrule
\end{tabular}
        \vspace{-2.5mm}
        \caption{The settings of panoramic representations.}\label{tab:supp_analysis}
    \end{minipage} \\
\end{table*}

\begin{figure*}[t]
    \centering
\includegraphics[width=.98\textwidth]{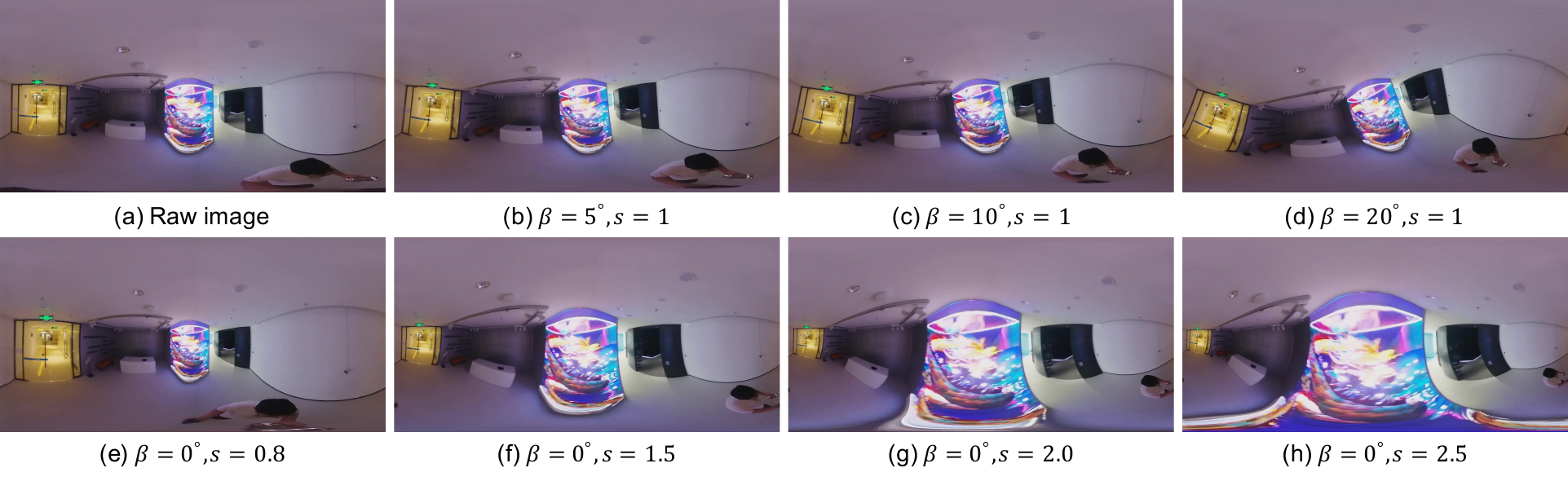}
\vspace{-5pt}
\caption{Visualization of panoramas under different transformations.}
\label{fig:mobius_sample}
\end{figure*}

\section{Metrics and Alignment}
\label{sec:metric}

\noindent \textbf{Metrics.} We evaluate with two standard metrics: Absolute Relative Error (\textit{AbsRel}) and Root Mean Squared Error (\textit{RMSE}). Performance assessments are confined to valid regions where ground truth depth, denoted as $D^{*}$, is available. We denote the number of valid pixels by $K$. Additionally, we employ three percentage metrics, $\delta_i$, for values of $i \in \{1.25, 1.25^2, 1.25^3\}$. With the predicted depth $D$, metrics can be formulated as follows:

\begin{itemize}
    \item Absolute Relative Error (\textit{AbsRel}):
    \begin{equation}
    \frac{1}{K}\sum_{i=1}^{K}\frac{||D(i) - D^{*}(i)||}{D^{*}(i)}.
    \end{equation}
    
    \item Root Mean Square Error (\textit{RMSE}):
    \begin{equation}
    \sqrt{\frac{1}{K}\sum_{i=1}^{K}|| D(i) - D^{*}(i)||^2}.
    \end{equation}

    \item $\delta_i$, the fraction of pixels where the relative error between the depth prediction $D$ and ground truth depth $D^{*}$ is less
than the threshold $i$:
     \begin{equation}
    \text{max}\{\frac{D(p)}{D^{*}(p)}, \frac{D^{*}(p)}{D(p)}\} < i.
    \end{equation}
\end{itemize}

\noindent \textbf{Alignment.} In the main paper, the reported results of PanDA-\{S,B,L\} in Tab.~\ref{tab:comparison_matterport3d},~\ref{tab:comparison_stanford2d3d} do not apply any alignment operation for a fair comparison. In addition, to assess the zero-shot performance of Depth Anything v1 and v2, Marigold, and our PanDA, we employ scale and shift alignment as described in~\cite{ranftl2020towards}. The scale and shift adjustments of the depth predictions are manually aligned with the depth ground truth. In Tab.~\ref{tab:analysis_erp}, ~\ref{tab:analysis_representation}, and Fig.~\ref{fig:mobius_plot}, this alignment is performed in the disparity space. Conversely, in Tab.~\ref{tab:comparison},~\ref{tab:ablation-supervision_loss},~\ref{tab:ablation-ssl_loss}, the alignment occurs in the depth space.

 \begin{figure}[t]
    \centering
\includegraphics[width=\linewidth]{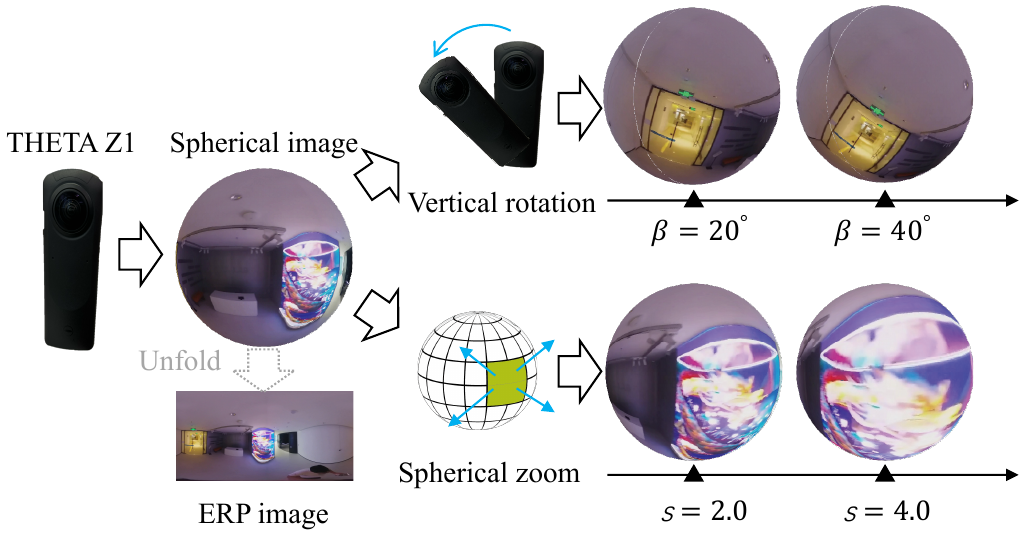}
\caption{Illustration of the spherical spatial transformations.}
\label{fig:mobius_visual}
\end{figure}

\section{Analysis}
\label{sec:supp_analysis}

\subsection{Different Panoramic Representations}
In Tab.~\ref{tab:supp_analysis}, we detail the settings of the panoramic representations used, including the number of patches, field-of-view (FoV), spatial resolution, and the grouping of the equator and polar regions. The indices of patches for CP, TP, and HS are illustrated in Fig.~\ref{fig:supp_index}.

\subsection{Different Camera Positions}

In Fig.~\ref{fig:benchmark_overview}, we utilize iPad Pro and the app "polycam" to scan and generate the point cloud of the scene.

\subsection{Various Spatial Transformations}

\noindent \textbf{Meaning.} As depicted in Fig.~\ref{fig:mobius_visual}, given the 360$^{\circ}$ camera, such as with the THETA Z1, it is not always possible to ensure that spherical images are captured vertically. Another scenario occurs in virtual reality (VR) environments, where users have the freedom to adjust their viewing directions and zoom in on objects of interest for an immersive experience. In these cases, spherical transformations are crucial to meet the practical demands of real-world applications.

  \begin{figure}[t]
    \centering
\includegraphics[width=\linewidth]{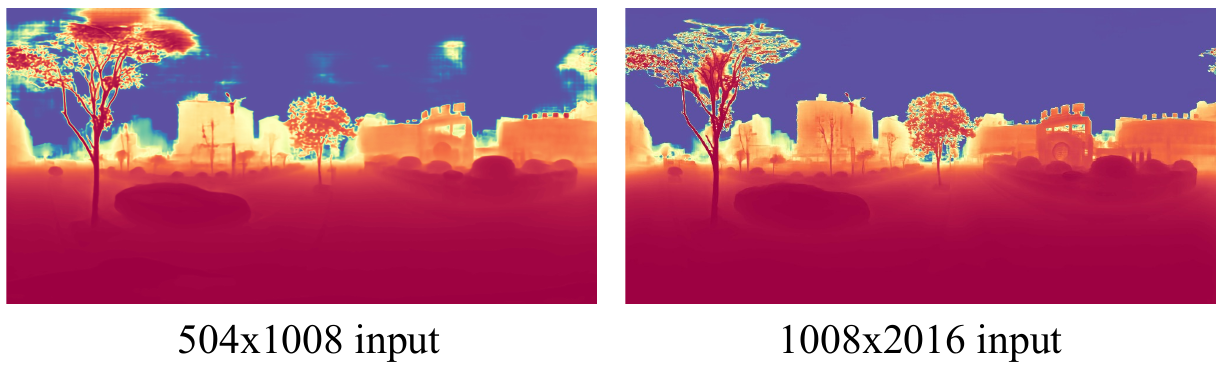}
\caption{Visual comparison between different input resolutions.}
\label{fig:supp_pseudo_resolution}
\end{figure}

\noindent \textbf{More visualization of spatial transformations.} Additional visualization results of the M\"obius transformation are presented in Fig.~\ref{fig:mobius_sample}, including vertical rotations with different angles $\beta$ and spherical zooms with different zoom levels $s$. It is obvious that the transformations introduce more curves, which complicates the task of panoramic depth estimation compared to panoramas captured vertically.

\section{The Proposed Method}
\label{sec:method}

\subsection{EPNL}

For each panorama, we sample 32 patches. The horizontal position of the patch center is randomly selected from a range of $0$ to $W$. For the vertical position, we use a Gaussian distribution to sample more patches around the equator region. The mean of this distribution is set at $\frac{H}{2}$, and its variance at $\frac{H}{6}$.

\subsection{Spatial Resolution of Pseudo Depth Labels}

As shown in Fig.~\ref{fig:supp_pseudo_resolution}, when generating pseudo depth labels for unlabeled panoramas, increasing the input resolution significantly reduces noise and enhances structural details.

\subsection{MTSA}

\begin{figure}[t!]
    \centering
\includegraphics[width=\linewidth]{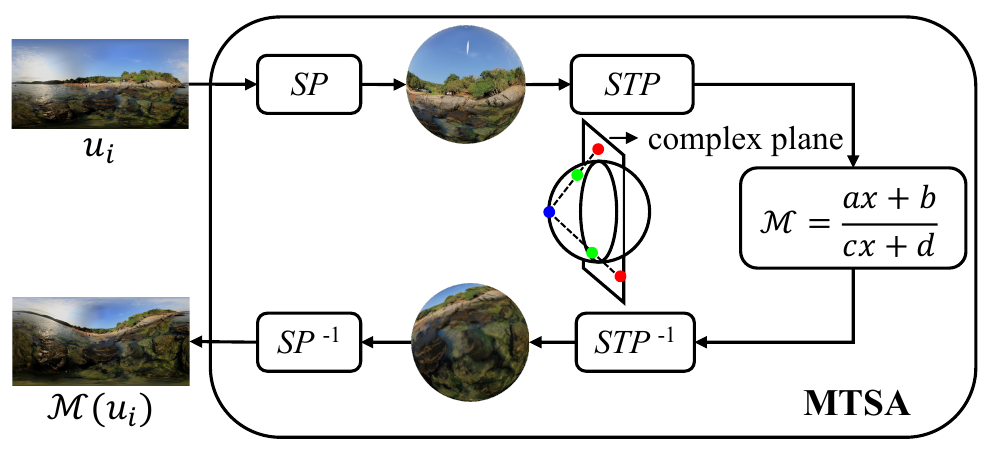}
\vspace{-15pt}
\caption{Illustration of the process of MTSA. \textcolor{blue}{$\bullet$}: Equator point; \textcolor{green}{$\bullet$}: Spherical point; \textcolor{red}{$\bullet$}: Complex plane point.}
\label{fig:MTSA}
\end{figure}

\textbf{Overview.} We illustrate the detailed process of the M\"obius transformation for panoramas. The formulas are based on~\cite{cao2023omnizoomer}. Differently, we take the equator center as the pole to zoom in on the objects at the equator. As illustrated in Fig.~\ref{fig:MTSA}, a panorama $u_i$ undergoes an initial projection from the plane to the sphere via spherical projection (SP). Subsequently, this spherical representation is projected onto the complex plane using stereographic projection (STP). In our conduction, the specific point on the complex plane is determined by the intersection of the equator point and a designated spherical point. The M\"obius transformation is applied on the complex plane. Following this, we apply the inverse stereographic projection (STP$^{-1}$) and inverse spherical projection (SP$^{-1}$) to obtain the transformed panorama $\mathcal{M}(u_i)$.

The M\"obius transformation is conducted in the complex plane. To achieve it, a panorama with ERP representation is first projected from the plane to the sphere via spherical projection (SP). The plane coordinate is proportional to the angle coordinate ($\theta, \phi$) (where $\theta$ represents the longitude and $\phi$ represents the latitude), while the spherical coordinate can be defined as ($x,y,z$). In this case, SP can be defined as follows~\cite{cao2023omnizoomer}:

\begin{equation}
\text{SP}: \begin{pmatrix}
     x  \\
     y \\
     z \\
\end{pmatrix} = 
\begin{pmatrix}
     \cos\phi\cos\theta \\
     \cos\phi\sin\theta \\
     \sin\phi \\
\end{pmatrix} .
\label{eq:sp}
\end{equation}

Then, we project from the sphere to the complex plane with stereographic projection (STP). By defining the coordinate of the complex plane as $Z=(x',y')$ and selecting the equator center as the pole, the STP can be formulated as follows:

\begin{equation}
\text{STP}: x' = {\frac {y}{1 - x}}\ ,\ y' = {\frac {z}{1 - x}}.
\label{eq:stp}
\end{equation}

In the complex plane, the M\"obius transformation is conducted with the following formulation:

{\begin{equation}
f(Z)={\frac {aZ+b}{cZ+d}}, 
\label{eq:mobius}
\end{equation}}

\noindent where $a$, $b$, $c$, and $d$ are complex numbers. In addition,  $a$, $b$, $c$, and $d$ should satisfy $ad-bc \neq 0$. For the vertical rotation with angle $\beta$, the parameters of M\"obius transformations can be represented as follows~\cite{cao2023omnizoomer}: 

\begin{equation}
\begin{pmatrix}
     a & b\\
     c & d \\
\end{pmatrix} = 
\begin{pmatrix}
     \cos\beta+j\sin\beta & 0\\
     0 & 1 \\
\end{pmatrix}.
\label{eq:rotation}
\end{equation}

For the zoom operation with level $s$, the parameters of M\"obius transformations can be represented as follows~\cite{cao2023omnizoomer}: 

\begin{equation}
\begin{pmatrix}
     a & b\\
     c & d \\
\end{pmatrix} = 
\begin{pmatrix}
    s & 0 \\
     0 & 1 \\
\end{pmatrix}.
\label{eq:zoom}
\end{equation}

The M\"obius transformation obeys the matrix chain multiplication rule. After the M\"obius transformation in the complex plane, we conduct inverse projections to project from the complex plane to the sphere and the plane, respectively. The inverse projections can be formulated as follows:

\begin{equation}
\begin{split}
    \text{STP}^{-1}: \begin{pmatrix}
         x  \\
         y \\
         z \\
    \end{pmatrix} &= 
    \begin{pmatrix}
         \frac{-1+x'^2+y'^2}{1+x'^2+y'^2} \\
         \frac{2x'}{1+x'^2+y'^2} \\
         \frac{2y'}{1+x'^2+y'^2} \\
        \end{pmatrix} \ ; \\
   \ \text{SP}^{-1}: \begin{pmatrix}
         \theta \\
         \phi \\
    \end{pmatrix} &= 
    \begin{pmatrix}
         \arctan(y/x) \\
         \arcsin(z) \\
        \end{pmatrix} \ .
\end{split}
    \label{eq:inverse}
\end{equation}

\section{More Experimental Results}
\label{sec:experiment}

\subsection{The parameters of MTSA}

For the proposed MTSA, the default setting is that: the vertical rotation angle is uniformly sampled in $[-10^{\circ},10^{\circ})$, denoted as $\mathcal{U}(-10^{\circ},10^{\circ})$. Moreover, the zoom level is uniformly sampled in $[1,1.5)$, denoted as $\mathcal{U}(1,1.5)$. To further discuss the effect of the MTSA by introducing vertical rotation and spherical zoom into spatial augmentation, we conduct ablation studies for the range of sampling distribution in Tab.~\ref{tab:supp_angle_range} and Tab.~\ref{tab:supp_zoom_range}.

\begin{table}[h]
    \begin{minipage}{\linewidth}
    \centering
        \tablestyle{1.5pt}{1.05}
        \begin{tabular}{y{27mm}|x{10mm}|x{9.5mm}x{9.5mm}|x{9.5mm}x{9.5mm}}
\toprule
\multirow{2}{*}{Methods} & \multirow{2}{*}{Original} & \multicolumn{2}{c|}{Vertical angle $\theta$} & \multicolumn{2}{c}{Zoom level $s$} \\
& & $10^{\circ}$ & $20^{\circ}$ & 2.0 & 3.0 \\
\midrule
$\mathcal{U}(-5^{\circ},5^{\circ})$ & \cellcolor[HTML]{efefef}\textbf{0.4896} & 0.5309 & 0.6045 & 0.6140 & 0.7426 \\
$\mathcal{U}(-10^{\circ},10^{\circ})$ & 0.4915 & 0.5188 & 0.5706 & 0.6242 & 0.7461 \\
$\mathcal{U}(-20^{\circ},20^{\circ})$ & 0.4923 & \cellcolor[HTML]{efefef}\textbf{0.5157} & 0.5500 & 0.5948 & 0.7205 \\
$\mathcal{U}(-30^{\circ},30^{\circ})$ & 0.5000 & 0.5208 & \cellcolor[HTML]{efefef}\textbf{0.5457} & \cellcolor[HTML]{efefef}\textbf{0.5939} & \cellcolor[HTML]{efefef}\textbf{0.7023} \\
\bottomrule
\end{tabular}
        \vspace{-2.5mm}
        \caption{Examine the range of vertical rotation angles. We report \textit{RMSE} metric on the Matterport3D dataset.}\label{tab:supp_angle_range}
        \vspace{-2mm}
    \end{minipage}
\end{table}

\begin{table}[h]
    \begin{minipage}{\linewidth}
    \centering
        \tablestyle{1.5pt}{1.05}
        \begin{tabular}{y{27mm}|x{10mm}|x{9.5mm}x{9.5mm}|x{9.5mm}x{9.5mm}}
\toprule
\multirow{2}{*}{Methods} & \multirow{2}{*}{Original} & \multicolumn{2}{c|}{Vertical angle $\theta$} & \multicolumn{2}{c}{Zoom level $s$} \\
& & $10^{\circ}$ & $20^{\circ}$ & 2.0 & 3.0 \\
\midrule
$\mathcal{U}(1,1.2)$ & 0.4943 & \cellcolor[HTML]{efefef}\textbf{0.5158} & \cellcolor[HTML]{efefef}\textbf{0.5667} & 0.7425 & 0.8870 \\
$\mathcal{U}(1,1.5)$ & \cellcolor[HTML]{efefef}\textbf{0.4915} & 0.5188 & 0.5706 & \cellcolor[HTML]{efefef}\textbf{0.6242} & \cellcolor[HTML]{efefef}\textbf{0.7461} \\
$\mathcal{U}(1,2)$ & 0.5250 & 0.5886 & 0.6890 & 0.8276 & 0.9649 \\
$\mathcal{U}(1,3)$ & 0.5187 & 0.5819 & 0.6841 & 0.8109 & 0.9494 \\
\bottomrule
\end{tabular}
        \vspace{-2.5mm}
        \caption{Examine the range of zoom levels. We report \textit{RMSE} metric on the Matterport3D dataset.}\label{tab:supp_zoom_range}
        \vspace{2mm}
    \end{minipage}
\end{table}

\noindent \textbf{Vertical rotation angle.} As shown in Tab.~\ref{tab:supp_angle_range}, it can be found that a smaller angle distribution can benefit the depth estimation of the original panorama. Moreover, MTSA with a larger angle distribution benefits the depth prediction on panoramas with larger rotation angles, \eg, 20$^{\circ}$, and larger zoom levels, \eg, 3.0. Transformations with larger vertical rotation angles and larger zoom levels would introduce severe curves to challenge the panoramic depth estimation. Our choice of $\mathcal{U}(-10^{\circ},10^{\circ})$ is a balance between the performance of original and transformed ones.

\noindent \textbf{Zoom level.} As depicted in Tab.~\ref{tab:supp_zoom_range}, we investigate the impact of various zoom level distributions. We observe that employing larger zoom level distributions, such as $\mathcal{U}(1,2)$ and $\mathcal{U}(1,3)$, can degrade the depth estimation performance for both original and transformed panoramas. We attribute this degradation to the severe distortions that hinder the model from learning effective structural information.

\subsection{Pseudo Depth Labels}

\noindent \textbf{Different amounts of pseudo depth labels.} To further examine the effect of pseudo depth labels in the SSL pipeline, we vary the amounts of pseudo depth labels, as illustrated in Tab.~\ref{tab:rebuttal_ratio}. The results show that the larger the amount of unlabeled data, the better the
performance, especially under spherical transformations. 

\begin{table}[h]
    \begin{minipage}{\linewidth}
    \centering
        \tablestyle{1.5pt}{1.05}
        \vspace{-2.5mm}
        \begin{tabular}{y{27mm}|x{10mm}|x{9.5mm}x{9.5mm}|x{9.5mm}x{9.5mm}}
\toprule
\multirow{2}{*}{Num. of unlabeled data} & \multirow{2}{*}{Original} & \multicolumn{2}{c|}{Vertical angle $\theta$} & \multicolumn{2}{c}{Zoom level $s$} \\
& & $10^{\circ}$ & $20^{\circ}$ & 2.0 & 3.0 \\
\midrule
10199 (10\%) & 0.4998 & 0.5280 & 0.5912 & 0.6892 & 0.8248 \\
20398 (20\%) & 0.4932 & 0.5252 & 0.5910 & 0.6919 & 0.8255 \\
101990 (100\%) & \cellcolor[HTML]{efefef}\textbf{0.4915} & \cellcolor[HTML]{efefef}\textbf{0.5188} & \cellcolor[HTML]{efefef}\textbf{0.5706} & \cellcolor[HTML]{efefef}\textbf{0.6242} & \cellcolor[HTML]{efefef}\textbf{0.7461} \\
\bottomrule
\end{tabular}
        \vspace{-3mm}
        \caption{Vary the number of unlabeled data during SSL. We report \textit{RMSE} metric on the Matterport3D dataset.}\label{tab:rebuttal_ratio}
        \vspace{-2mm}
    \end{minipage}
\end{table}

\noindent \textbf{Only pseudo depth labels for training.} To further investigate the effect of pseudo depth labels from the teacher model, in Tab.~\ref{tab:supp_only_pseudo}, we only utilize the unlabeled panoramas and the corresponding pseudo depth labels to train the student model. It is observed that training with only pseudo depth labels yields better performance compared to solely using synthetic depth ground truth. This improvement is likely due to several factors: 1) The amount of pseudo depth labels exceeds that of synthetic depth ground truth; 2) The unlabeled data consists of real-world samples; 3) The teacher model provides accurate pseudo labels that enhance student model training. However, both approaches show limited effectiveness in transformed panoramas.

\begin{table}[h]
    \begin{minipage}{\linewidth}
    \centering
        \tablestyle{1.5pt}{1.05}
        \begin{tabular}{y{27mm}|x{10mm}|x{9.5mm}x{9.5mm}|x{9.5mm}x{9.5mm}}
\toprule
\multirow{2}{*}{Methods} & \multirow{2}{*}{Original} & \multicolumn{2}{c|}{Vertical angle $\theta$} & \multicolumn{2}{c}{Zoom level $s$} \\
& & $10^{\circ}$ & $20^{\circ}$ & 2.0 & 3.0 \\
\midrule
$\mathcal{L}_{\text{S}}$ & 0.5109 & 0.5711 & 0.6804 & 0.8381 & \cellcolor[HTML]{efefef}\textbf{0.9793} \\
$\mathcal{L}_{\text{P}}$ & \cellcolor[HTML]{efefef}\textbf{0.5031} & \cellcolor[HTML]{efefef}\textbf{0.5584} & \cellcolor[HTML]{efefef}\textbf{0.6557} & \cellcolor[HTML]{efefef}\textbf{0.8358} & 0.9870 \\
\bottomrule
\end{tabular}
        \vspace{-2.5mm}
        \caption{The effect of only utilizing the pseudo depth labels to train the student model. We report \textit{RMSE} metric on the Matterport3D dataset.}\label{tab:supp_only_pseudo}
        \vspace{2mm}
    \end{minipage}
\end{table}

\subsection{Few-shot Learning for Fine-Tuning}

The student model has been trained using both synthetic data and large-scale unlabeled data. We explore whether a small amount of real-world panoramic depth ground truth is sufficient to fine-tune our PanDA for real-world scenes. In this context, Tab.~\ref{tab:supp_few_shot} demonstrates the results from uniformly sampling the Matterport3D dataset~\cite{chang2017matterport3d} at percentages of 1\%, 5\%, 10\%, and 25\%. It is observed that with just 5\% of the samples, our PanDA can be fine-tuned to achieve competitive results with existing SOTA panoramic monocular depth estimation methods. Additionally, at 25\%, the performance closely approximates that achieved by using all the depth ground truth in the training set of Matterport3D~\cite{chang2017matterport3d}.

\begin{table}[t]
    \begin{minipage}{\linewidth}
    \centering
        \tablestyle{1.5pt}{1.05}
        \begin{tabular}{y{26mm}|x{11mm}x{11mm}|x{9mm}x{9mm}x{9mm}}
\toprule
Percentage & \textit{AbsRel} $\downarrow$ & \textit{RMSE} $\downarrow$ & $\delta_1$ $\uparrow$ & $\delta_2$ $\uparrow$ & $\delta_3$ $\uparrow$ \\
\midrule
1\% & 0.1340 & 0.5303 & 83.26 & 96.01 & 98.94 \\
5\% & 0.1099 & 0.4356 & 89.48 & 97.93 & 99.31 \\
10\% & 0.1002 & 0.4236 & 90.77 & 98.03 & 99.40 \\
25\% & 0.0946 & 0.3967 & 91.91 & 98.26 & 99.47 \\
100\% & \cellcolor[HTML]{efefef}\textbf{0.0922} & \cellcolor[HTML]{efefef}\textbf{0.3950} & \cellcolor[HTML]{efefef}\textbf{92.26 }& \cellcolor[HTML]{efefef}\textbf{98.30} & \cellcolor[HTML]{efefef}\textbf{99.47} \\
\bottomrule
\end{tabular}
        \vspace{-2.5mm}
        \caption{Utilizing small parts of the training set of the Matterport dataset for fine-tuning.}\label{tab:supp_few_shot}
    \end{minipage}
\end{table}

\subsection{LoRA Rank}

By default, the rank parameter in LoRA is set as 4. In Tab.~\ref{tab:supp_lora_rank}, it can be found that different choices of the rank parameter have a limited effect on the depth estimation performance.

\begin{table}[h]
    \begin{minipage}{\linewidth}
    \centering
        \tablestyle{1.5pt}{1.05}
        \begin{tabular}{y{27mm}|x{12mm}x{12mm}|x{12mm}x{12mm}}
\toprule
Rank & 2 & 4 & 8 \\
\midrule
\textit{AbsRel} $\downarrow$ & 0.1049 & \cellcolor[HTML]{efefef}\textbf{0.1036} & 0.1047 \\
\textit{RMSE} $\downarrow$ & \cellcolor[HTML]{efefef}\textbf{0.4531} & 0.4539 & 0.4583 \\
\bottomrule
\end{tabular}
        \vspace{-2.5mm}
        \caption{The effect of LoRA rank parameter. We report \textit{RMSE} metric on the Matterport3D dataset.}\label{tab:supp_lora_rank}
    \end{minipage}
\end{table}

\subsection{The effect of Sampling Regions in EPNL}

In Tab.~\ref{tab:rebuttal_sampling}, by changing the sampling regions from equator region to polar regions, the performance degrades. We ascribe it as the polar regions contain less structural information. Sampling on the polar regions provides less structural guidance.

\begin{table}[h]
    \begin{minipage}{\linewidth}
    \centering
        \tablestyle{1.5pt}{0.9}
        \begin{tabular}{y{27mm}|x{12mm}x{12mm}|x{12mm}x{12mm}}
\toprule
\multirow{2}{*}{Methods} & \multicolumn{2}{c}{Matterport3D} & \multicolumn{2}{c}{Stanford2D3D} \\
& \textit{AbsRel} $\downarrow$ & \textit{RMSE} $\downarrow$ & \textit{AbsRel} $\downarrow$ & \textit{RMSE} $\downarrow$ \\
\midrule
Sampling in Poles & 0.1489 & 0.5403 & 0.1274 & 0.3542 \\
Sampling in Equator & \cellcolor[HTML]{efefef}\textbf{0.1256} & \cellcolor[HTML]{efefef}\textbf{0.5062} & \cellcolor[HTML]{efefef}\textbf{0.1109} & \cellcolor[HTML]{efefef}\textbf{0.3401} \\
\bottomrule
\end{tabular}
        \vspace{-2.5mm}
        \caption{Change to poles (Latitude [-90$^\circ$,-30$^\circ$] \textbf{$\cup$} [30$^\circ$,90$^\circ$]).}\label{tab:rebuttal_sampling}
    \end{minipage}
\end{table}

\subsection{The visualization Issue of DAMs}

\begin{figure}[h]
    \centering
\includegraphics[width=\linewidth]{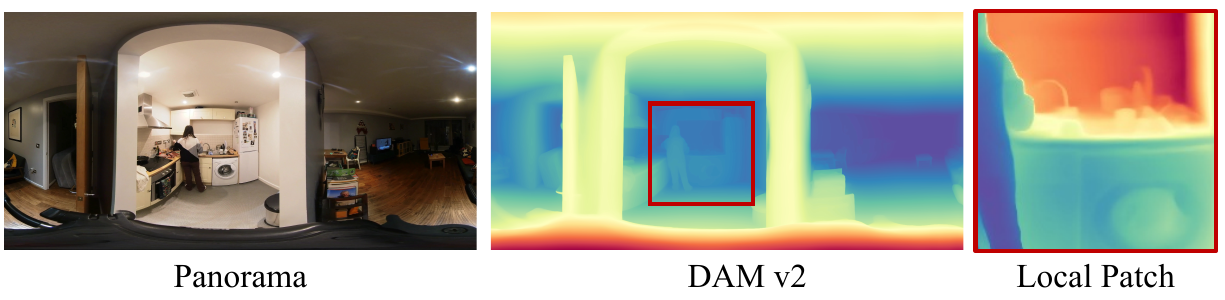}
\vspace{-15pt}
\caption{Illustration of the visualization issue of DAMs.}
\label{fig:supp_visual_illustrate}
\end{figure}

As illustrated in Fig.~\ref{fig:supp_visual_illustrate}, some structural details can be neglected if we visualize the depth estimation result of DAM as a whole. This is because the global normalization before visualization would squeeze the details of local regions. Therefore, for a fair comparison, we showcase the local areas of the DAM prediction with local normalization.

\subsection{Point Cloud Results}

\begin{figure}[h]
    \centering
\includegraphics[width=\linewidth]{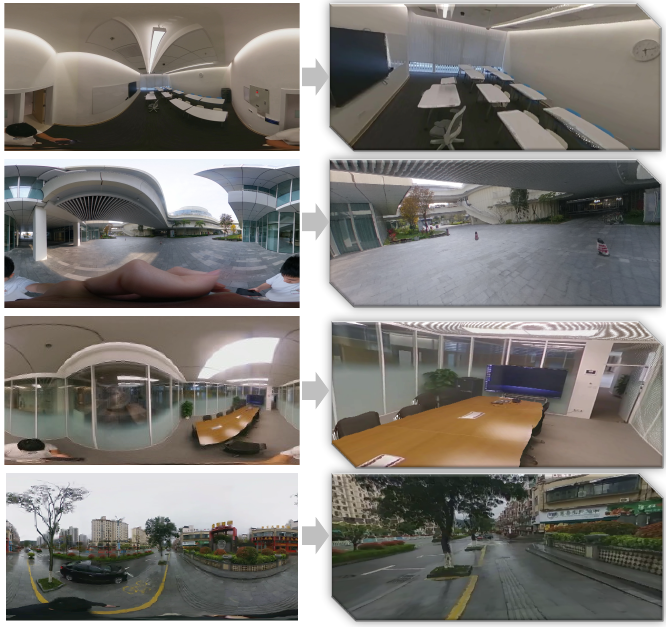}
\vspace{-15pt}
\caption{Visualization of point clouds generated from the depth estimation results of our PanDA.}
\label{fig:supp_point_cloud}
\end{figure}

In Fig.~\ref{fig:supp_point_cloud}, the point clouds generated from our depth predictions can recover reasonable structures of the scene, such as the chairs in the classroom and outdoor buildings.

\subsection{Model Complexity}

With only LoRA added, the parameters of PanDA are similar to DAM v2. As for inference speed, processing a 504$\times$1008 panorama requires 49/90/234ms with PanDA-\{S,B,L\}, respectively. The running speeds are tested by averaging 100 times on an A40 GPU.

\section{Limitation and Future Work}

Due to the scarcity of panoramic depth labels in diverse scenes, our teacher model is trained on limited scenes compared with the depth datasets for perspective images. To enhance the zero-shot capability of our model, future work will focus on collecting panoramas paired with depth labels across a broader range of environments, including both synthetic and real-world scenes.
{
    \small
    \bibliographystyle{ieeenat_fullname}
    \bibliography{main}
}

\end{document}